\documentclass[runningheads]{llncs}
\usepackage{graphicx}
\graphicspath{{./images/}}  % No need to prepend images/ to the path.
\usepackage{tikz}
\usepackage{comment}
\usepackage{amsmath,amssymb} % define this before the line numbering.
\usepackage{color}
\usepackage{orcidlink}

\usepackage{pgfplots}
\usepgfplotslibrary{groupplots}
\pgfplotsset{compat=1.17}

\usepackage{booktabs}
\usepackage{multirow}
\usepackage{array}
\usepackage[caption=false]{subfig}

\usepackage[accsupp]{axessibility}  % Improves PDF readability for those with disabilities.

\usepackage{xspace}
\makeatletter
\DeclareRobustCommand\onedot{\futurelet\@let@token\@onedot}
\newcommand\@onedot{\ifx\@let@token.\else.\null\fi\xspace}
\makeatother

\newcommand\eg{\emph{e.g}\onedot} 
\newcommand\ie{\emph{i.e}\onedot} 
 
\newcommand\etc{\emph{etc}\onedot} 
\newcommand\wrt{w.r.t\onedot} 
 
\newcommand\etal{\emph{et al}\onedot}

\usepackage[capitalize,nameinlink]{cleveref}
\Crefname{section}{Sec.}{Secs.}
\Crefname{section}{Section}{Sections}
\Crefname{table}{Tab.}{Tabs.}
\Crefname{table}{Table}{Tables}

\begin{document}
% \linenumbers
\pagestyle{headings}
\mainmatter
\def\ECCVSubNumber{556}  % Insert your submission number here

\title{\texorpdfstring{Scene Text Recognition with\\Permuted Autoregressive Sequence Models}{Scene Text Recognition with Permuted Autoregressive Sequence Models}} % Replace with your title

\begin{comment}
\titlerunning{ECCV-22 submission ID \ECCVSubNumber} 
\authorrunning{ECCV-22 submission ID \ECCVSubNumber} 
\author{Anonymous ECCV submission}
\institute{Paper ID \ECCVSubNumber}
\end{comment}
%******************

% \begin{comment}
\titlerunning{Scene Text Recognition with Permuted Autoregressive Sequence Models}
\author{Darwin Bautista\orcidlink{0000-0002-1050-0776} \and Rowel Atienza\orcidlink{0000-0002-8830-2534}}
\authorrunning{D. Bautista \and R. Atienza}
\institute{Electrical and Electronics Engineering Institute,\\
University of the Philippines, Diliman\\
\email{\{darwin.bautista,rowel\}@eee.upd.edu.ph}}
% \end{comment}
%******************
\maketitle

\begin{abstract}
Context-aware STR methods typically use internal autoregressive (AR) language models (LM). Inherent limitations of AR models motivated two-stage methods which employ an external LM. The conditional independence of the external LM on the input image may cause it to erroneously rectify correct predictions, leading to significant inefficiencies. Our method, PARSeq, learns an ensemble of internal AR LMs with shared weights using Permutation Language Modeling. It unifies context-free non-AR and context-aware AR inference, and iterative refinement using bidirectional context. Using synthetic training data, PARSeq achieves state-of-the-art (SOTA) results in STR benchmarks (91.9\% accuracy) and more challenging datasets. It establishes new SOTA results (96.0\% accuracy) when trained on real data. PARSeq is optimal on accuracy vs parameter count, FLOPS, and latency because of its simple, unified structure and parallel token processing. Due to its extensive use of attention, it is robust on arbitrarily-oriented text which is common in real-world images. Code, pretrained weights, and data are available at: 
\url{https://github.com/baudm/parseq}.

\end{abstract}

\section{Introduction}

Machines read text in natural scenes by first detecting text regions, then recognizing text in those regions. The task of recognizing text from the cropped regions is called Scene Text Recognition (STR). STR enables reading of road signs, billboards, paper bills, product labels, logos, printed shirts, \etc. It has practical applications in self-driving cars, augmented reality, retail, education, and devices for the visually-impaired, among others. In contrast to Optical Character Recognition (OCR) in documents where the text attributes are more uniform, STR has to deal with varying font styles, orientations, text shapes, illumination, amount of occlusion, and inconsistent sensor conditions. Images captured in natural environments could also be noisy, blurry, or distorted. In essence, STR is an important but very challenging problem.

STR is mainly a vision task, but in cases where parts of the text are impossible to read, \eg due to an occluder, the image features alone will not be enough to make accurate inferences. In such cases, language semantics is typically used to aid the recognition process. Context-aware STR methods incorporate semantic priors from a word representation model \cite{Qiao_2020_CVPR} or dictionary \cite{Nguyen_2021_CVPR}, or learned from data \cite{shi2016robust,Lee_2016_CVPR,Baek_2019_ICCV,sheng2019nrtr,lee2020recognizing,yu2020towards,Fang_2021_CVPR,shi2018aster,bleeker2020bidirectional} using sequence modeling \cite{DBLP:journals/corr/BahdanauCB14,vaswani2017attention}.

Sequence modeling has the advantage of learning end-to-end trainable language models (LM). STR methods with \textit{internal} LMs jointly process image features and language context. They are trained by enforcing an autoregressive (AR) constraint on the language context where \textit{future} tokens are conditioned on \textit{past} tokens but not the other way around, resulting in the model $P(\mathbf{y} | \mathbf{x}) = \prod^T_{t=1} P(y_t | \mathbf{y}_{<t}, \mathbf{x})$ where $\mathbf{y}$ is the $T$-length text label of the image $\mathbf{x}$. AR models have two inherent limitations arising from this constraint. First, the model is able to learn the token dependencies in one direction only---usually the left-to-right (LTR) direction. This unidirectionality causes AR models to be biased towards a single reading direction, resulting in spurious addition of suffixes or direction-dependent predictions (illustrated in Appendix A). Second, during inference, the AR model can only be used to output tokens serially in the same direction used for training. This is called next-token or monotonic AR decoding.

\begin{figure}
  \centering
    \subfloat[ABINet]{%
         \includegraphics[width=0.48\textwidth]{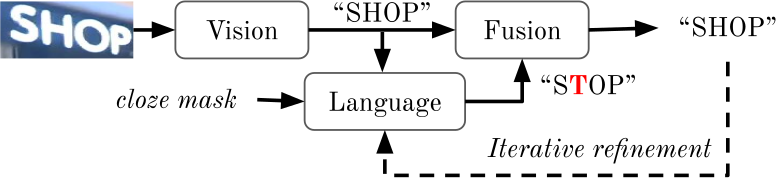}%
         \label{fig:parseq-vs-sota:abinet}%
     }
    \hfill
    \subfloat[Unified STR model (Ours)]{%
         \includegraphics[width=0.48\textwidth]{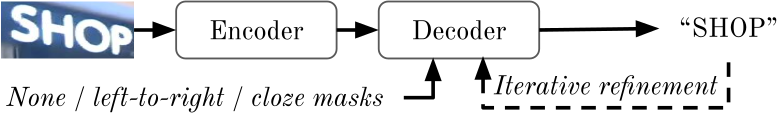}%
         \label{fig:parseq-vs-sota:ours}%
     }
   \caption{\protect\subref{fig:parseq-vs-sota:abinet} State-of-the-art method ABINet \cite{Fang_2021_CVPR} uses a combination of context-free vision and context-aware language models. The language model functions as a \textit{spell checker} but is prone to erroneous rectification of correct initial predictions due to its conditional independence on the image features. \protect\subref{fig:parseq-vs-sota:ours} Our proposed method performs both initial decoding and iterative refinement by jointly processing image and context features, resulting in a single holistic output. This eschews the need for separate language and fusion models resulting in a more efficient and robust STR method}
   \label{fig:parseq-vs-sota}
\end{figure}

To address these limitations, prior works have combined left-to-right and right-to-left (RTL) AR models \cite{shi2018aster,bleeker2020bidirectional}, or opted for a two-stage approach using an ensemble of a context-free STR model with a standalone or external LM \cite{yu2020towards,Fang_2021_CVPR}. A combined LTR and RTL AR model still suffers from unidirectional context but works around it by performing two separate decoding streams---one for each direction---then choosing the prediction with the higher likelihood. Naturally, this results in increased decoding time and complexity. Meanwhile, two-stage ensemble approaches like in \Cref{fig:parseq-vs-sota:abinet} obtain their initial predictions using parallel non-AR decoding. The initial context-less prediction is decoded directly from the image using the context-free model $P(\mathbf{y} | \mathbf{x}) = \prod^T_{t=1} P(y_t | \mathbf{x})$. This enables the external LM, $P(\mathbf{y}) = \prod^T_{t=1} P(y_t | \mathbf{y}_{\neq t})$ in ABINet \cite{Fang_2021_CVPR} for example, to use bidirectional context since all characters are available at once. The LM functions as a \textit{spell checker} and rectifies the initial prediction, producing a context-based output. The conditional independence of the LM from the input image may cause it to erroneously rectify correct predictions if they appear misspelled, or if a similar word with a higher likelihood exists. This is evident in the low word accuracy of the LM in SRN (27.6\%) and in ABINet (41.9\%) when used as a spell checker \cite{Fang_2021_CVPR}. Hence, a separate fusion layer is used to combine the features from the initial prediction and the LM prediction to get the final output. A closer look at the LM of ABINet (Appendix B) reveals that it is inefficient for STR. It is underutilized relative to its parameter count, and it exhibits dismal word accuracy despite using a significant chunk of the overall compute requirements of the full ABINet model.

In sequence model literature, there has been recent interest in generalized models of sequence generation. Various neural sequence models, such as AR and refinement-based non-AR, were shown to be special cases in the generalized framework proposed by Mansimov \etal \cite{mansimov2019generalized}. This result posits that the same generalization can be done in STR models, unifying context-free and context-aware STR. While the advantages of this unification are not apparent, we shall show later that such a generalized model enables the use of an internal LM while maintaining the refinement capabilities of an external LM.

\begin{figure}
  \centering
 \includegraphics[width=0.84\textwidth]{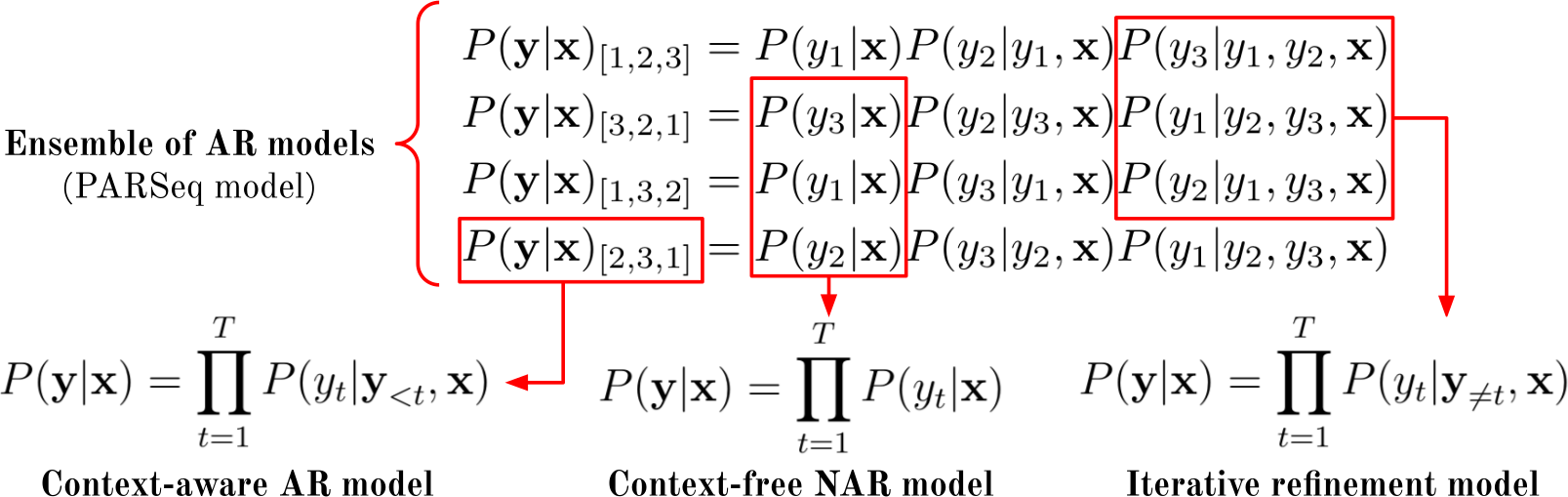}
   \caption{Illustration of NAR and iterative refinement (cloze) models in relation to an ensemble of AR models for an image $\mathbf{x}$ with a three-element text label $\mathbf{y}$. Four different factorizations of $P(\mathbf{y} | \mathbf{x})$ (out of six possible) are shown, with each one determined by the factorization order shown in the subscript}
   \label{fig:contexts-example}
\end{figure}

Permutation Language Modeling (PLM) was originally proposed for large-scale language pretraining \cite{yang2019xlnet}, but recent works \cite{tian2020train,qi2021bang} have adapted it for learning Transformer-based generalized sequence models capable of different decoding schemes. In this work, we adapt PLM for STR. PLM can be considered a generalization of AR modeling, and a PLM-trained model can be seen as an ensemble of AR models with shared architecture and weights \cite{uria2014deep}. With the use of attention masks for dynamically specifying token dependencies, such a model, illustrated in \Cref{fig:contexts-example}, can learn and use conditional character probabilities given an arbitrary subset of the input context, enabling monotonic AR decoding, parallel non-AR decoding, and even iterative refinement.

In summary, state-of-the-art (SOTA) STR methods \cite{yu2020towards,Fang_2021_CVPR} opted for a two-stage ensemble approach in order to use bidirectional language context. The low word accuracy of their external LMs, despite increased training and runtime requirements, highlights the need for a more efficient approach. To this end, we propose a \textbf{p}ermuted \textbf{a}uto\textbf{r}egressive \textbf{seq}uence (PARSeq) model for STR. Trained with PLM, PARSeq is a unified STR model with a simple structure, but is capable of both context-free and context-aware inference, as well as iterative refinement using bidirectional (\textit{cloze}) context. PARSeq achieves SOTA results on the STR benchmarks for both synthetic and real training data (\Cref{tab:main-results}) across all character sets (\Cref{tab:charset-results}), while being optimal in its use of parameters, FLOPS, and runtime (\Cref{fig:compute-requirements}). For a more comprehensive comparison, we also benchmark on larger and more difficult real datasets which contain occluded and arbitrarily-oriented text (\Cref{fig:art-coco-uber-samples}). PARSeq likewise achieves SOTA results in these datasets (\Cref{tab:new-results}).

\section{Related Work}

The recent surveys of Long \etal \cite{long2021scene} and Chen \etal \cite{chen2021text} provide comprehensive discussions on different approaches in STR. In this section, we focus on the use of language semantics in STR.

\textbf{Context-free STR} methods directly predict the characters from image features. The output characters are conditionally-independent of each other. The most prominent approaches are CTC-based \cite{graves2006connectionist} methods \cite{shi2016end,liu2016star,wang2017gated,borisyuk2018rosetta}, with a few using different approaches such as self-attention \cite{dosovitskiy2020image} for pooling features into character positions \cite{atienza2021vitstr}, or casting STR as a multi-instance classification problem \cite{Jaderberg14c,https://doi.org/10.48550/arxiv.2102.10884}. Ensemble methods \cite{yu2020towards,Fang_2021_CVPR} use an attention mechanism \cite{DBLP:journals/corr/BahdanauCB14,vaswani2017attention} to produce the initial context-less predictions. Since context-free methods rely solely on the image features for prediction, they are less robust against corruptions like occluded or incomplete characters. This limitation motivated the use of language semantics for making the recognition model more robust.

\textbf{Context-aware STR} methods typically use semantics learned from data to aid in recognition. Most approaches \cite{Baek_2019_ICCV,Lee_2016_CVPR,shi2016robust,cheng2017focusing,shi2018aster} use RNNs with attention \cite{DBLP:journals/corr/BahdanauCB14} or Transformers \cite{sheng2019nrtr,lee2020recognizing,bleeker2020bidirectional} to learn internal LMs using the standard AR training. These methods are limited to monotonic AR decoding. Ensemble methods \cite{yu2020towards,Fang_2021_CVPR} use bidirectional context via an external LM for prediction refinement. The conditional independence of the external LM on image features makes it prone to erroneous rectification, limiting usefulness while incurring significant overhead. VisionLAN \cite{Wang_2021_ICCV_visionlan} learns semantics by selectively masking image features of individual characters during training, akin to denoising autoencoders and Masked Language Modeling (MLM) \cite{devlin-etal-2019-bert}. In contrast to prior work, PARSeq learns an internal LM using PLM instead of the standard AR modeling. It supports flexible decoding by using a parameterization which decouples the target decoding position from the input context, similar to the \textit{query stream} of two-stream attention \cite{yang2019xlnet}. Unlike ABINet \cite{Fang_2021_CVPR} which uses the \textit{cloze} context for both training and inference, PARSeq uses it for iterative refinement only. Moreover, as said earlier, the refinement model of ABINet is conditionally independent of the input image, while PARSeq considers both input image and language context in the refinement process.

\textbf{Generation from Sequence Models} can be categorized into two contrasting schemes: autoregressive (one token at a time) and non-autoregressive (all tokens predicted at once). Mansimov \etal \cite{mansimov2019generalized} proposed a generalized framework for sequence generation which unifies the said schemes. BANG \cite{qi2021bang} adapted two-stream attention \cite{yang2019xlnet} for use with MLM, in contrast to our use of PLM. PMLM \cite{liao2020probabilistically} is trained using a generalization of MLM where the masking ratio is stochastic. A variant which uses a uniform prior was shown to be equivalent to a PLM-trained model. Closest to our work is Tian \etal \cite{tian2020train} which adapts the two-stream attention parameterization \cite{yang2019xlnet} to decoders by interspersing the content and query streams from different layers. In contrast, our decoder does not use self-attention and does not intersperse the two streams. This allows our single layer decoder to use the query stream only, and avoid the overhead of the unused content stream.

\section{Permuted Autoregressive Sequence Models}
\label{ch:methodology}

In this section, we first present the Transformer-based model architecture of PARSeq. Next, we discuss how to train it using Permutation Language Modeling. Lastly, we show how to use the trained model for inference by discussing the different decoding schemes and the iterative refinement procedure.

\subsection{Model Architecture}

Multi-head Attention (MHA) \cite{vaswani2017attention} is extensively used by PARSeq. We denote it as $MHA(\mathbf{q}, \mathbf{k}, \mathbf{v}, \mathbf{m})$, where $\mathbf{q}$, $\mathbf{k}$, and $\mathbf{v}$ refer to the required parameters \textit{query}, \textit{key}, and \textit{value}, while $\mathbf{m}$ refers to the optional attention mask. We provide the background material on MHA in Appendix C.

\begin{figure*}
  \centering
  \includegraphics[width=\linewidth]{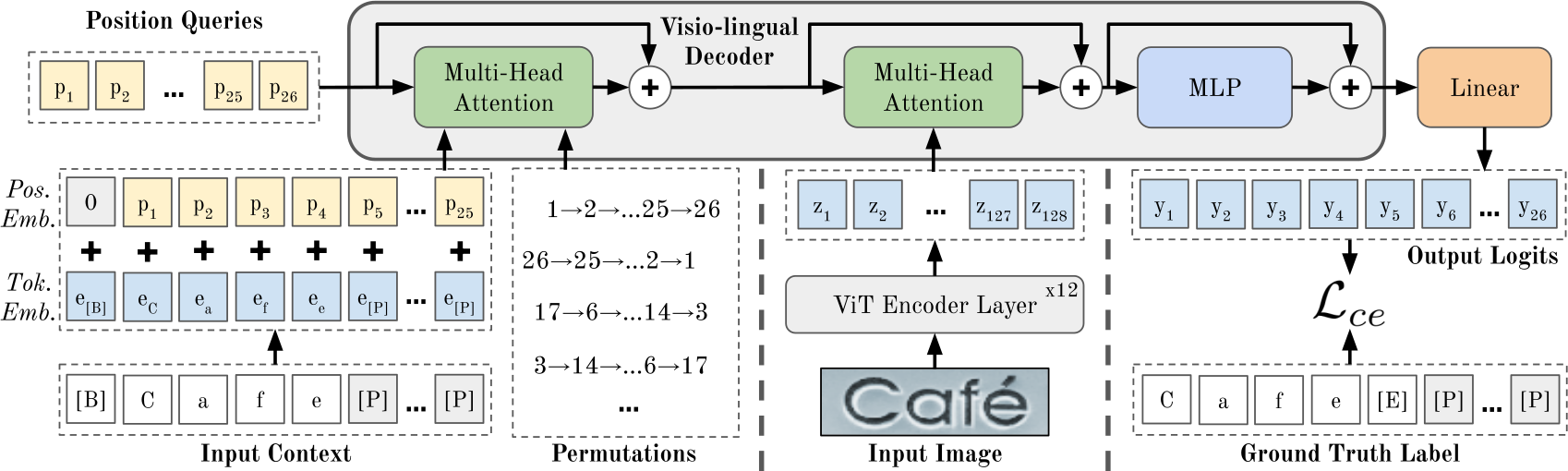}
   \caption[PARSeq architecture and training overview.]{PARSeq architecture and training overview. $LayerNorm$ and $Dropout$ layers are omitted due to space constraints. \texttt{[B]}, \texttt{[E]}, and \texttt{[P]} stand for \textit{beginning-of-sequence (BOS)}, \textit{end-of-sequence (EOS)}, and \textit{padding} tokens, respectively. $T = 25$ results in 26 distinct \textit{position} tokens. The position tokens both serve as query vectors and position embeddings for the input context. For \texttt{[B]}, no position embedding is added. Attention masks are generated from the given permutations and are used only for the \textit{context-position} attention. $\mathcal{L}_{ce}$ pertains to the cross-entropy loss}
   \label{fig:system-overview}
\end{figure*}

PARSeq follows an encoder-decoder architecture, shown in \Cref{fig:system-overview}, commonly used in sequence modeling tasks. The encoder has 12 layers while the decoder is only a single layer. This \textit{deep-shallow} configuration \cite{kasai2021deep} is a deliberate design choice which minimizes the overall computational requirements of the model while having a negligible impact in performance. Details in Appendix D.

\subsubsection{ViT Encoder.} Vision Transformer (ViT) \cite{dosovitskiy2020image} is the direct extension of the Transformer to images. A ViT layer contains one MHA module used for \textit{self-attention}, \ie $\mathbf{q} = \mathbf{k} = \mathbf{v}$. The encoder is a 12-layer ViT without the classification head and the \texttt{[CLS]} token. An image $\mathbf{x} \in \mathbb{R}^{W\times{H}\times{C}}$, with width $W$, height $H$, and number of channels $C$, is \textit{tokenized} by evenly dividing it into $p_w\times{p_h}$ patches, flattening each patch, then linearly projecting them into $d_{model}$-dimensional tokens using a patch embedding matrix $\mathbf{W}^p \in \mathbb{R}^{p_w p_h C \times d_{model}}$, resulting in $(WH) / (p_w p_h)$ tokens. Learned position embeddings of equal dimension are added to the tokens prior to being processed by the first ViT layer. In contrast to the standard ViT, all output tokens $\mathbf{z}$ are used as input to the decoder:

\begin{equation}
    \mathbf{z} = Enc(\mathbf{x}) \in \mathbb{R}^{\frac{WH}{p_w p_h} \times d_{model}}
  \label{eq:encoder-output}
\end{equation}

\subsubsection{Visio-lingual Decoder.} The decoder follows the same architecture as the pre-$LayerNorm$ \cite{baevski2018adaptive,wang2019learning} Transformer decoder but uses twice the number of attention heads, \ie $nhead = d_{model} / 32$. It has three required inputs consisting of \textit{position}, \textit{context}, and \textit{image} tokens, and an optional attention mask.

In the following equations, we omit $LayerNorm$ and $Dropout$ for brevity. The first \textit{MHA} module is used for \textit{context--position} attention:

\begin{equation}
    \mathbf{h}_c = \mathbf{p} + MHA(\mathbf{p}, \mathbf{c}, \mathbf{c}, \mathbf{m}) \in \mathbb{R}^{(T + 1) \times d_{model}}
  \label{eq:mha-context-position}
\end{equation} where $T$ is the context length, $\mathbf{p} \in \mathbb{R}^{(T + 1) \times d_{model}}$ are the position tokens, $\mathbf{c} \in \mathbb{R}^{(T + 1) \times d_{model}}$ are the context embeddings with positional information, and $\mathbf{m} \in \mathbb{R}^{(T + 1) \times (T + 1)}$ is the optional \textit{attention mask}. Note that the use of special \textit{delimiter} tokens (\texttt{[B]} or \texttt{[E]}) increases the total sequence length to $T + 1$.

The \textit{position} tokens encode the target position to be predicted, each one having a direct correspondence to a specific position in the output. This parameterization is similar to the query stream of two-stream attention \cite{yang2019xlnet}. It decouples the context from the target position, allowing the model to learn from PLM. Without the position tokens, \ie if the context tokens are used as \textit{queries} themselves like in standard Transformers, the model will not learn anything meaningful from PLM and will simply function like a \textit{standard} AR model.

The supplied mask varies depending on how the model is used. During training, masks are generated from random permutations (\Cref{sec:plm}). At inference (\Cref{sec:decoding-schemes}), it could be a standard left-to-right lookahead mask (AR decoding), a \textit{cloze} mask (iterative refinement), or no mask at all (NAR decoding).

The second MHA is used for \textit{image--position} attention:

\begin{equation}
    \mathbf{h}_i = \mathbf{h}_c + MHA(\mathbf{h}_c, \mathbf{z}, \mathbf{z}) \in \mathbb{R}^{(T + 1) \times d_{model}}
  \label{eq:mha-image-position}
\end{equation} where no attention mask is used. The last decoder hidden state is the output of the MLP, $\mathbf{h}_{dec} = \mathbf{h}_i + MLP(\mathbf{h}_i) \in \mathbb{R}^{(T + 1) \times d_{model}}$.

Finally, the output logits are $\mathbf{y} = Linear(\mathbf{h}_{dec}) \in \mathbb{R}^{(T + 1) \times (S + 1)}$ where $S$ is the size of the character set (charset) used for training. The additional character pertains to the \texttt{[E]} token (which marks the end of the sequence). In summary, given an attention mask $\mathbf{m}$, the decoder is a function which takes the form:
\begin{equation}
    \mathbf{y} = Dec(\mathbf{z}, \mathbf{p}, \mathbf{c}, \mathbf{m}) \in \mathbb{R}^{(T + 1) \times (S + 1)}
  \label{eq:dec}
\end{equation} 

\subsection{Permutation Language Modeling}
\label{sec:plm}

Given an image $\mathbf{x}$, we want to maximize the likelihood of its text label $\mathbf{y} = [y_1, y_2, \dots, y_T]$ under the set of model parameters $\theta$. In standard AR modeling, the likelihood is factorized using the chain rule according to the canonical ordering, $[1, 2, \dots, T]$, resulting in the model $\log p(\mathbf{y} | \mathbf{x}) = \sum_{t=1}^T{\log p_\theta(y_{t} | \mathbf{y}_{<t}, \mathbf{x})}$. However, Transformers process all tokens in parallel, allowing the output tokens to \textit{access} or be conditionally-dependent on all the input tokens. In order to have a valid AR model, \textit{past} tokens cannot have access to \textit{future} tokens. The AR property is enforced in Transformers with the use of attention masks. For example, a standard AR model for a three-element sequence $\mathbf{y}$ will have the attention mask shown in \Cref{tab:ar-mask}.

The key idea behind PLM is to train on all $T!$ factorizations of the likelihood:

\begin{equation}
    \label{eqn:plm-objective}
    \log p(\mathbf{y} | \mathbf{x}) = \mathbb{E}_{\mathbf{z} \sim \mathcal{Z}_T} \left[ \sum_{t=1}^T{\log p_\theta(y_{z_t} | \mathbf{y}_{\mathbf{z}_{<t}}, \mathbf{x})} \right]
\end{equation} where $\mathcal{Z}_T$ denotes the set of all possible permutations of the index sequence [1, 2, \dots, $T$], and $z_t$ and $\mathbf{z}_{<t}$ denote the $t$-th element and the first $t - 1$ elements, respectively, of a permutation $\mathbf{z} \in \mathcal{Z}_T$. Each permutation $\mathbf{z}$ specifies an ordering which corresponds to a distinct factorization of the likelihood.

To implement PLM in Transformers, we do \textbf{not} need to actually permute the text label $\mathbf{y}$. Rather, we craft the attention mask to \textit{enforce} the ordering specified by $\mathbf{z}$. As a concrete example, shown in \Cref{tab:attention-masks} are attention masks for four different permutations of a three-element sequence. Notice that while the order of the input and output sequences remains constant, all four correspond to distinct AR models specified by the given permutation or factorization order. With this in mind, it can be seen that the standard AR training is just a special case of PLM where only one permutation, $[1,2,\dots, T]$, is used.

In practice, we cannot train on all $T!$ factorizations due to the exponential increase in computational requirements. As a compromise, we only use $K$ of the possible $T!$ permutations. Instead of sampling uniformly, we choose the $K$ permutations in a specific way. We use $K / 2$ permutation pairs. The first half consists of the \textit{left-to-right} permutation, [1, 2, \dots, $T$], and $K / 2 - 1$ randomly sampled permutations. The other half consists of \textit{flipped} versions of the first. We found that this sampling procedure results in a more stable training.

With $K$ permutations and the ground truth label $\hat{\mathbf{y}}$, the full training loss is the mean of the individual cross-entropy losses for each permutation-derived attention mask $\mathbf{m}_k$:

\begin{equation}
    \mathcal{L} = \frac{1}{K}\sum_{k=1}^K{\mathcal{L}_{ce}(\mathbf{y}_k, \hat{\mathbf{y}})}
  \label{eq:char-logits}
\end{equation} where $\mathbf{y}_k = Dec(\mathbf{z}, \mathbf{p}, \mathbf{c}, \mathbf{m}_k)$. \textit{Padding} tokens are ignored in the loss computation. More PLM details are in Appendix E.

\begin{table}
\setlength\tabcolsep{1pt}
\newcolumntype{C}[1]{>{\centering\arraybackslash}p{#1}}
\caption{Illustration of AR attention masks for each permutation. The table header (with the \texttt{[B]} token) pertains to the input context, while the header column (with the \texttt{[E]} token) corresponds to the output tokens. \textit{1} means that the output token has conditional dependency on the corresponding input token. \textit{0} means that no information flows from input to output}
\label{tab:attention-masks}
\captionsetup{position=top}
\subfloat[{$[1,2,3]$}]{%
\centering
  \begin{tabular}{ c | C{12pt} C{12pt} C{12pt} C{12pt} }
    & \texttt{[B]} & $y_1$ & $y_2$ & $y_3$ \\
    \midrule
    $y_1$ & 1 & 0 & 0 & 0 \\
    $y_2$ & 1 & 1 & 0 & 0 \\
    $y_3$ & 1 & 1 & 1 & 0 \\
    \texttt{[E]} & 1 & 1 & 1 & 1 \\
  \end{tabular}%
  \label{tab:ar-mask}%
}
\hfill
\subfloat[{$[3,2,1]$}]{%
\centering
  \begin{tabular}{ c | C{12pt} C{12pt} C{12pt} C{12pt} }
    & \texttt{[B]} & $y_1$ & $y_2$ & $y_3$ \\
    \midrule
    $y_1$ & 1 & 0 & 1 & 1 \\
    $y_2$ & 1 & 0 & 0 & 1 \\
    $y_3$ & 1 & 0 & 0 & 0 \\
    \texttt{[E]} & 1 & 1 & 1 & 1 \\
  \end{tabular}%
}
\hfill
\subfloat[{$[1,3,2]$}]{%
\centering
  \begin{tabular}{ c | C{12pt} C{12pt} C{12pt} C{12pt} }
    & \texttt{[B]} & $y_1$ & $y_2$ & $y_3$ \\
    \midrule
    $y_1$ & 1 & 0 & 0 & 0 \\
    $y_2$ & 1 & 1 & 0 & 1 \\
    $y_3$ & 1 & 1 & 0 & 0 \\
    \texttt{[E]} & 1 & 1 & 1 & 1 \\
  \end{tabular}%
}
\hfill
\subfloat[{$[2,3,1]$}]{%
\centering
  \begin{tabular}{ c | C{12pt} C{12pt} C{12pt} C{12pt} }
    & \texttt{[B]} & $y_1$ & $y_2$ & $y_3$ \\
    \midrule
    $y_1$ & 1 & 0 & 1 & 1 \\
    $y_2$ & 1 & 0 & 0 & 0 \\
    $y_3$ & 1 & 0 & 1 & 0 \\
    \texttt{[E]} & 1 & 1 & 1 & 1 \\
  \end{tabular}%
}
\end{table}

\subsection{Decoding Schemes}
\label{sec:decoding-schemes}

PLM training coupled with the correct parameterization allows PARSeq to be used with various decoding schemes. In this work, we only use two contrasting schemes even though more are theoretically supported. Specifically, we elaborate the use of monotonic AR and NAR decoding, as well as iterative refinement.

% Semi-autoregressive (SAR) decoding pertains to a combination of AR and NAR. The decoding process is iterative but multiple tokens are generated at a time. SAR is theoretically supported by PARSeq but we do not explore it in this work.

\textbf{Autoregressive (AR)} decoding generates one new token per iteration. The \textit{left-to-right} attention mask (\Cref{tab:ltr-ar-mask}) is always used. For the first iteration, the context is set to \texttt{[B]}, and only the first position query token $\mathbf{p}_1$ is used. For any succeeding iteration $i$, position queries $[\mathbf{p}_1, \dots, \mathbf{p}_i]$ are used, while the context is set to the previous output, $argmax(\mathbf{y})$ prepended with \texttt{[B]}.

\textbf{Non-autoregressive (NAR)} decoding generates all output tokens at the same time. All position queries $[\mathbf{p}_1, \dots, \mathbf{p}_{T + 1}]$ are used but no attention mask is used (\Cref{tab:nar-mask}). The context is always \texttt{[B]}.

\textbf{Iterative refinement} can be performed regardless of the initial decoding method (AR or NAR). The previous output (truncated at \texttt{[E]}) serves as the context for the current iteration similar to AR decoding, but all position queries $[\mathbf{p}_1, \dots, \mathbf{p}_{T + 1}]$ are always used. The \textit{cloze} attention mask (\Cref{tab:cloze-mask}) is used. It is created by starting with an all-one mask, then masking out the matching token positions.

\begin{table}
\setlength\tabcolsep{2pt}
\newcolumntype{C}[1]{>{\centering\arraybackslash}p{#1}}
\caption{Illustration of information flow for the different decoding schemes. Conventions follow \Cref{tab:attention-masks}. In NAR decoding, no mask is used; this is equivalent to using an all-one mask. "\dots" pertains to elements $y_3$ to $y_{T-1}$}
\captionsetup{position=top}
\subfloat[\textit{left-to-right} AR mask]{%
\centering
  \begin{tabular}{ c | C{12pt} C{12pt} C{12pt} C{12pt} C{12pt} }
    & \texttt{[B]} & $y_1$ & $y_2$ & \dots & $y_T$ \\
    \midrule
    $y_1$ & 1 & 0 & 0 & 0 & 0 \\
    $y_2$ & 1 & 1 & 0 & 0 & 0 \\
    \dots & 1 & 1 & 1 & \dots & 0 \\
    $y_T$ & 1 & 1 & 1 & 1 & 0 \\
    \texttt{[E]} & 1 & 1 & 1 & 1 & 1 \\
  \end{tabular}%
  \label{tab:ltr-ar-mask}
}
\hfill
\subfloat[NAR mask]{\parbox{0.2\textwidth}{%
\centering
  \begin{tabular}{ c | c }
    & \texttt{[B]} \\
    \midrule
    $y_1$ & 1 \\
    $y_2$ & 1 \\
    \dots & 1 \\
    $y_T$ & 1 \\
    \texttt{[E]} & 1 \\
  \end{tabular}%
  \label{tab:nar-mask}%
}}
\hfill
\subfloat[\textit{cloze} mask]{%
\centering
  \begin{tabular}{ c | C{12pt} C{12pt} C{12pt} C{12pt} C{12pt} }
    & \texttt{[B]} & $y_1$ & $y_2$ & \dots & $y_T$ \\
    \midrule
    $y_1$ & 1 & 0 & 1 & 1 & 1 \\
    $y_2$ & 1 & 1 & 0 & 1 & 1 \\
    \dots & 1 & 1 & 1 & \dots & 1 \\
    $y_T$ & 1 & 1 & 1 & 1 & 0 \\
    \texttt{[E]} & 1 & 1 & 1 & 1 & 1 \\
  \end{tabular}%
  \label{tab:cloze-mask}%
}
\end{table}

\section{Results and Analysis}
\label{ch:results-analysis}

In this section, we first discuss the experimental setup including the datasets, preprocessing methods, training and evaluation protocols, and metrics used. Next, we present our results and compare PARSeq to SOTA methods in terms of the said metrics and commonly used computational cost indicators.

\subsection{Datasets}

STR models are traditionally trained on large-scale synthetic datasets because of the relative scarcity of labelled real data \cite{Baek_2019_ICCV}. However, in recent years, the amount of labelled real data has become sufficient for training STR models. In fact, training on real data was shown to be more sample-efficient than on synthetic data \cite{Baek_2021_CVPR}. Hence, in addition to the commonly used synthetic training datasets MJSynth (MJ) \cite{Jaderberg14c} and SynthText (ST) \cite{Gupta16}, we also use real data for training. Specifically, we use COCO-Text (COCO)  \cite{veit2016cocotext}, RCTW17 \cite{shi2017icdar2017}, Uber-Text (Uber) \cite{UberText}, ArT \cite{chng2019icdar2019}, LSVT \cite{sun2019icdar}, MLT19 \cite{nayef2019icdar2019}, and ReCTS \cite{zhang2019icdar}. A comprehensive discussion about these datasets is available in Baek \etal \cite{Baek_2021_CVPR}. In addition, we also use two recent large-scale real datasets based on Open Images \cite{OpenImages2}: TextOCR \cite{singh2021textocr} and annotations from the OpenVINO toolkit \cite{pmlr-v157-krylov21a}. More details in Appendix F.

Following prior works \cite{Baek_2019_ICCV}, we use IIIT 5k-word (IIIT5k) \cite{mishra2012scene}, CUTE80 (CUTE) \cite{risnumawan2014robust}, Street View Text (SVT) \cite{wang2011end}, SVT-Perspective (SVTP) \cite{phan2013recognizing}, ICDAR 2013 (IC13) \cite{karatzas2013icdar}, and ICDAR 2015 (IC15) \cite{karatzas2015icdar} as the datasets for evaluation. Baek \etal \cite{Baek_2019_ICCV} provides an in-depth discussion of these datasets. We use the case-sensitive annotations of Long and Yao \cite{long2020unreal} for IIIT5k, CUTE, SVT, and SVTP. Note that IC13 and IC15 have two \textit{versions} of their respective \textit{test} splits commonly used in the literature---857 and 1,015 for IC13; 1,811 and 2,077 for IC15. To avoid confusion, we refer to the \textit{benchmark} as the union of IIIT5k, CUTE, SVT, SVTP, IC13 \textit{(1,015)}, and IC15 \textit{(2,077)}.

\begin{figure}
    \centering
    \subfloat[Samples from the \textit{benchmark} datasets]{%
        \includegraphics[height=18mm]{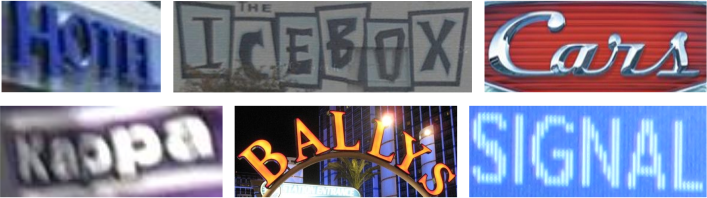}%
        \label{fig:benchmark-samples}%
    }
    \hfill
    \subfloat[Samples from Uber, COCO, ArT]{%
        \includegraphics[height=18mm]{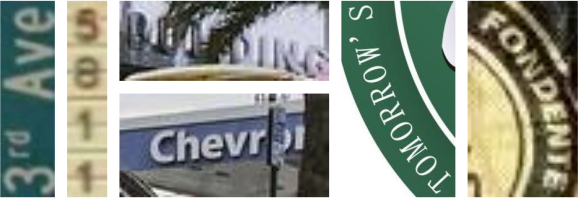}%
        \label{fig:art-coco-uber-samples}%
    }
    \caption{Sample test images from the datasets used}
\end{figure}

These six benchmark datasets only have a total of 7,672 test samples. This amount pales in comparison to benchmark datasets used in other vision tasks such as ImageNet \cite{imagenet_cvpr09} (\textit{classification}, 50k samples) and COCO \cite{lin2014microsoft} (\textit{detection}, 40k samples). Furthermore, the said datasets largely contain horizontal text only, as shown in \Cref{fig:benchmark-samples}, except for SVT, SVTP, and IC15 \textit{2,077} which contain a number of rotated text. In the real world, the conditions are less ideal, and captured text will most likely be blurry, vertically-oriented, rotated, or even occluded. In order to have a more comprehensive comparison, we also use the test sets of more recent datasets, shown in \Cref{fig:art-coco-uber-samples}, such as COCO-Text (9.8k samples; low-resolution, occluded text), ArT \cite{chng2019icdar2019} (35.1k samples; curved and rotated text), and Uber-Text \cite{UberText} (80.6k samples; vertical and rotated text).

\subsection{Training Protocol and Model Selection}
\label{sec:training}

All models are trained in a mixed-precision, dual-GPU setup using PyTorch DDP for 169,680 iterations with a batch size of 384. Learning rates vary per model (Appendix G.2). The Adam \cite{kingma:adam} optimizer is used together with the 1cycle \cite{smith2019super} learning rate scheduler. At iteration 127,260 (75\% of total), Stochastic Weight Averaging (SWA) \cite{Izmailov_swa_2018} is used and the 1cycle scheduler is replaced by the SWA scheduler. Validation is performed every 1,000 training steps. Since SWA averages weights at the end of each epoch, the last checkpoint at the end of training is selected. For PARSeq, $K = 6$ permutations are used (\Cref{sec:num-perms-vs-acc}). A patch size of $8\times{4}$ is used for PARSeq and ViTSTR. More details are in Appendix G.

\textbf{Label preprocessing} is done following prior work \cite{shi2018aster}. For training, we set a maximum label length of $T = 25$, and use a charset of size $S = 94$ which contains mixed-case alphanumeric characters and punctuation marks.

\textbf{Image preprocessing} is done like so: images are first augmented, resized, then finally normalized to the interval $[-1, 1]$. The set of augmentation operations consists primarily of RandAugment \cite{cubuk2020randaugment} operations, excluding \texttt{Sharpness}. \texttt{Invert} is added due to its effectiveness in house number data \cite{autoaugment}. \texttt{GaussianBlur} and \texttt{PoissonNoise} are also used due to their effectiveness in STR data augmentation \cite{atienza2021data}. A RandAugment policy with 3 layers and a magnitude of 5 is used. Images are resized unconditionally to 128$\times$32 pixels.

\subsection{Evaluation Protocol and Metrics}

All experiments are performed on an NVIDIA Tesla A100 GPU system. Reported mean$\pm$SD values are obtained from four replicates per model. A t-test ($\alpha$ = 0.05) is used to determine if model differences are statistically-significant. There can be multiple \textit{best} results in a column if the differences are not statistically-significant. PARSeq results are obtained from the \textbf{same} model using two different decoding schemes: PARSeq$_A$ denotes AR decoding with one refinement iteration, while PARSeq$_N$ denotes NAR decoding with two refinement iterations (ablation study in Appendix H).

\textbf{Word accuracy} is the primary metric for STR benchmarks. A prediction is considered correct if and only if characters at all positions match.

\textbf{Charset} may vary at inference time. Subsets of the training charset can be used for evaluation. Specifically, the following charsets are used: 36-character (lowercase alphanumeric), 62-character (mixed-case alphanumeric), and 94-char\-acter (mixed-case alphanumeric with punctuation). In Python, these correspond to array slices \texttt{[:36]}, \texttt{[:62]}, and \texttt{[:94]} of \texttt{string.printable}, respectively.

\subsection{Ablation on training permutations vs test accuracy}
\label{sec:num-perms-vs-acc}

As discussed in \Cref{sec:plm}, training on all possible permutations is not feasible in practice due to the exponential increase in computational requirements. We instead sample a number of permutations from the pool of all possible permutations. \Cref{tab:num-perms} shows the effect of the number of training permutations on the test accuracy for all decoding schemes. With $K = 1$, only the left-to-right ordering is used and the training simplifies to the standard AR modeling. In this setup, NAR decoding does not work at all, while AR decoding works well as expected. Meanwhile, the refinement or \textit{cloze} accuracy is at a dismal 71.14\% (this is very low considering that the ground truth itself is used as the initial prediction). All decoding schemes start to perform satisfactorily only at $K >= 6$. This result shows that PLM is indeed required to achieve a unified STR model. Intuitively, NAR decoding will not work when training on just the forward and/or reverse orderings ($K <= 2$) because the variety of training contexts is insufficient. NAR decoding relies on the priors for each character which could only be sufficiently trained if all characters in the charset naturally exist as the first character of a sequence. Ultimately, $K = 6$ provides the best balance between decoding accuracy and training time. The very high cloze accuracy ($\sim$94\%) of our internal LM highlights the advantage of jointly using image features and language context for prediction refinement. After all, the primary input signal in STR is the image, not the language context.

\begin{table}
  \centering
  \setlength\tabcolsep{5pt}
  \caption{94-char word accuracy on the benchmark vs number of permutations ($K$) used for training PARSeq. No refinement iterations were used for both AR and NAR decoding. \textit{cloze acc.} pertains to the word accuracy of one refinement iteration. It was measured by using the ground truth label as the initial prediction}
  \begin{tabular}{ r | r r r r }
    \toprule
    $K$ & AR acc. & NAR acc. & cloze acc. & Training hours \\
    \midrule
    1 & 93.04 & 0.01 & 71.14 & 5.86 \\
    2 & 93.48 & 22.69 & 94.55 & 7.30 \\
    6 & 93.34 & 92.22 & 94.81 & 8.48 \\
    12 & 92.91 & 91.71 & 94.59 & 10.10 \\
    24 & 92.67 & 91.72 & 94.36 & 13.53 \\
    \bottomrule
  \end{tabular}
  \label{tab:num-perms}
\end{table}

\subsection{Comparison to state-of-the-art (SOTA)}

We compare PARSeq to popular and recent SOTA methods. In addition to the published results, we reproduce a select number of methods for a fair comparison \cite{Baek_2019_ICCV}. In \Cref{tab:main-results}, most reproduced methods attain higher accuracy compared to the original results. The exception is ABINet (around 1.4\% decline in combined accuracy) which originally used a much longer training schedule (with pretraining of 80 and 8 epochs for LM and VM, respectively) and additional data (WikiText-103). For both synthetic and real data, PARSeq$_A$ achieves the highest word accuracies, while PARSeq$_N$ consistently places second or third. When real data is used, all reproduced models attain much higher accuracy compared to the original reported results, while PARSeq$_A$ establishes new SOTA results.

In \Cref{tab:charset-results}, we show the mean accuracy for each charset. When synthetic data is used for training, there is a steep decline in accuracy from the 36- to the 62- and 94-charsets. This suggests that diversity of cased characters is lacking in the synthetic datasets. Meanwhile, PARSeq$_A$ consistently achieves the highest accuracy on all charset sizes. Finally in \Cref{tab:new-results}, PARSeq is the most robust against occlusion and text orientation variability. Appendix J contains more experiments on arbitrarily-oriented text. Notice that the accuracy gap between methods is better revealed by these larger and more challenging datasets.

\begin{table*}
  \centering
  \scriptsize
  \setlength{\tabcolsep}{5pt}
  \caption{Mean word accuracy on the benchmark vs evaluation charset size}
  \begin{tabular}{ c c c c c }
    \toprule
    Method & Train data & 36-char & 62-char & 94-char \\
    \midrule
    CRNN & S & 83.2$\pm$0.2 & 56.5$\pm$0.3 & 54.8$\pm$0.2 \\
    ViTSTR-S & S & 88.6$\pm$0.0 & 69.5$\pm$1.0 & 67.7$\pm$1.0 \\
    TRBA & S & 90.6$\pm$0.1 & 71.9$\pm$0.9 & 69.9$\pm$0.8 \\
    ABINet & S & 89.8$\pm$0.2 & 68.5$\pm$1.1 & 66.4$\pm$1.0 \\
    PARSeq$_{N}$ & S & 90.7$\pm$0.2 & 72.5$\pm$1.1 & 70.5$\pm$1.1 \\
    PARSeq$_{A}$ & S & \textbf{91.9}$\pm$0.2 & \textbf{75.5}$\pm$0.6 & \textbf{73.0}$\pm$0.7 \\
    \midrule
    CRNN & R & 88.5$\pm$0.1 & 87.2$\pm$0.1 & 85.8$\pm$0.1 \\
    ViTSTR-S & R & 94.3$\pm$0.1 & 92.8$\pm$0.1 & 91.8$\pm$0.1 \\
    TRBA & R & 95.2$\pm$0.2 & 93.7$\pm$0.1 & 92.5$\pm$0.1 \\
    ABINet & R & 95.2$\pm$0.1 & 93.7$\pm$0.1 & 92.4$\pm$0.1 \\
    PARSeq$_{N}$ & R & 95.2$\pm$0.1 & 93.7$\pm$0.1 & 92.7$\pm$0.1 \\
    PARSeq$_{A}$ & R & \textbf{96.0}$\pm$0.0 &
    \textbf{94.6}$\pm$0.0 & \textbf{93.3}$\pm$0.1 \\
    \bottomrule
  \end{tabular}
  \label{tab:charset-results}
\end{table*}

\begin{table*}
  \scriptsize
  \centering
  \setlength{\tabcolsep}{5pt}
  \caption{36-char word accuracy on larger and more challenging datasets}
  \begin{tabular}{ c c c c c | c }
    \toprule
     & & \multicolumn{4}{c}{Test datasets and \# of samples} \\
    \cmidrule{3-6}
    \multirow{2}{*}{Method} & Train & ArT & COCO & Uber & Total \\
    & data & 35,149 & 9,825 & 80,551 & 125,525 \\
    \midrule
    CRNN & S & 57.3$\pm$0.1 & 49.3$\pm$0.6 & 33.1$\pm$0.3 & 41.1$\pm$0.3 \\
    ViTSTR-S & S & 66.1$\pm$0.1 & 56.4$\pm$0.5 & 37.6$\pm$0.3 & 47.0$\pm$0.2 \\
    TRBA & S & 68.2$\pm$0.1 & 61.4$\pm$0.4 & 38.0$\pm$0.3 & 48.3$\pm$0.2 \\
    ABINet & S & 65.4$\pm$0.4 & 57.1$\pm$0.8 & 34.9$\pm$0.3 & 45.2$\pm$0.3 \\
    PARSeq$_{N}$ & S & 69.1$\pm$0.2 & 60.2$\pm$0.8 & 39.9$\pm$0.5 & 49.7$\pm$0.3 \\
    PARSeq$_{A}$ & S & \textbf{70.7}$\pm$0.1 & \textbf{64.0}$\pm$0.9 & \textbf{42.0}$\pm$0.5 & \textbf{51.8}$\pm$0.4 \\
    \midrule
    CRNN & R & 66.8$\pm$0.2 & 62.2$\pm$0.3 & 51.0$\pm$0.2 & 56.3$\pm$0.2 \\
    ViTSTR-S & R & 81.1$\pm$0.1 & 74.1$\pm$0.4 & 78.2$\pm$0.1 & 78.7$\pm$0.1 \\
    TRBA & R & 82.5$\pm$0.2 & 77.5$\pm$0.2 & 81.2$\pm$0.3 & 81.3$\pm$0.2 \\
    ABINet & R & 81.2$\pm$0.1 & 76.4$\pm$0.1 & 71.5$\pm$0.7 & 74.6$\pm$0.4 \\
    PARSeq$_{N}$ & R & 83.0$\pm$0.2 & 77.0$\pm$0.2 & 82.4$\pm$0.3 & 82.1$\pm$0.2 \\
    PARSeq$_{A}$ & R & \textbf{84.5}$\pm$0.1 & \textbf{79.8}$\pm$0.1 & \textbf{84.5}$\pm$0.1 & \textbf{84.1}$\pm$0.0 \\
    \bottomrule
  \end{tabular}
  \label{tab:new-results}
\end{table*}

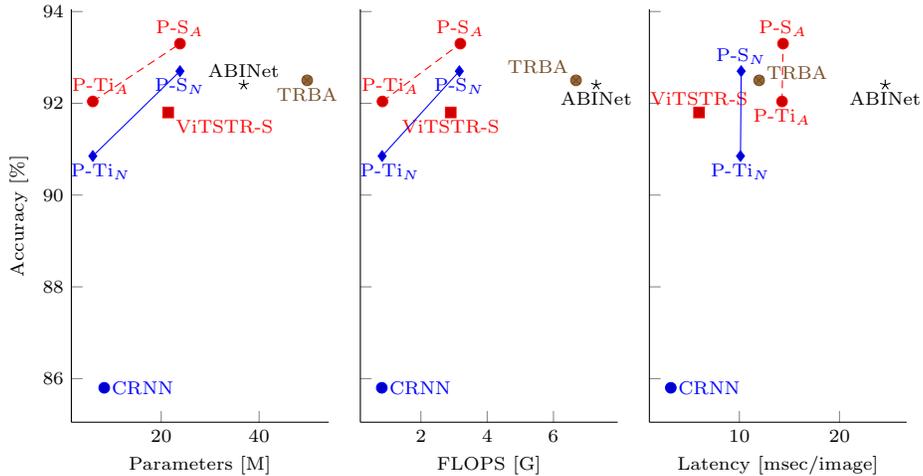
\begin{figure*}
%   \footnotesize
  \scriptsize
  \centering
    \begin{tikzpicture}[auto]
    \begin{groupplot}[
        group style={
            % set how the plots should be organized
            group size=3 by 1,
            % only show ticklabels and axis labels on the bottom
            y descriptions at=edge left,
            % set the `vertical sep' to zero
            %vertical sep=0pt,
            horizontal sep=12pt,
        },
        clip = false,
        ylabel={Accuracy [\%]},
        axis lines* = left,
        width=0.41\linewidth,
        height=0.58\linewidth,
        ]
        
    \nextgroupplot[
        xlabel={Parameters [M]},
        cycle list name=color
    ]
        \addplot coordinates { (8.36,85.8) } node [right] {CRNN};
        \addplot coordinates { (21.42,91.8) } node [below right] {ViTSTR-S};
        \addplot coordinates { (49.82,92.5) } node [below] {TRBA};
        \addplot coordinates { (36.86,92.4) } node [above] {ABINet};
        \addplot coordinates { (6.02,90.85)
        (23.83,92.7)}
            node[right=3pt,below, pos = 0] {P-Ti$_{N}$}
            node[below, pos = 1] {P-S$_{N}$};
        \addplot coordinates { (6.02,92.04) (23.83,93.3)}
            node[right=3pt,above, pos = 0] {P-Ti$_{A}$}
            node[above, pos = 1] {P-S$_{A}$};;

    \nextgroupplot[
        xlabel={FLOPS [G]},
        cycle list name=color
    ]
        \addplot coordinates { (0.82,85.8) } node [right] {CRNN};
        \addplot coordinates { (2.90,91.8) } node [below] {ViTSTR-S};
        \addplot coordinates { (6.68,92.5) } node [above left] {TRBA};
        \addplot coordinates { (7.29,92.4) } node [below] {ABINet};
        \addplot coordinates { (0.83,90.85)
        (3.16,92.7)}
            node[right=3pt,below, pos = 0] {P-Ti$_{N}$}
            node[below, pos = 1] {P-S$_{N}$};
        \addplot coordinates { (0.84,92.04) (3.19,93.3)}
            node[right=3pt,above, pos = 0] {P-Ti$_{A}$}
            node[above, pos = 1] {P-S$_{A}$};
        
    \nextgroupplot[
        xlabel={Latency [msec/image]},
        cycle list name=color
    ]
        \addplot coordinates { (3.15,85.8) } node[right] {CRNN};
        \addplot coordinates { (5.97,91.8) } node[above] {ViTSTR-S};
        \addplot coordinates { (11.97,92.5) } node[above=3pt,right] {TRBA};
        \addplot coordinates { (24.57,92.4) } node[below] {ABINet};
        \addplot coordinates { (10.11,90.85)
        (10.18,92.7)}
            node[below, pos = 0] {P-Ti$_{N}$}
            node[above, pos = 1] {P-S$_{N}$};
        \addplot coordinates { (14.26,92.04) (14.36,93.3)}
            node[below, pos = 0] {P-Ti$_{A}$}
            node[above, pos = 1] {P-S$_{A}$};
    
    % \legend{CRNN,ViTSTR-S,TRBA,ABINet,PARSeq$_N$,PARSeq$_A$}
    
    \end{groupplot}
    \end{tikzpicture}

   \caption{Mean word accuracy (94-char) vs computational cost. \textit{P-S} and \textit{P-Ti} are shorthands for PARSeq-S and PARSeq-Ti, respectively. For TRBA and PARSeq$_{A}$, FLOPS and latency correspond to mean values measured on the benchmark}
   \label{fig:compute-requirements}
\end{figure*}

\Cref{fig:compute-requirements} shows the cost-quality trade-offs in terms of accuracy and commonly used cost indicators like parameter count, FLOPS, and latency. PARSeq-S is the base model used for all results, while -Ti is its scaled down variant (details in Appendix D). Note that for PARSeq, the parameter count is fixed regardless of the decoding scheme. PARSeq-S achieves the highest mean word accuracy and exhibits very competitive cost--quality characteristics across the three indicators. Compared to ABINet and TRBA, PARSeq-S uses significantly less parameters and FLOPS. In terms of latency (Appendix I), PARSeq-S with AR decoding is slightly slower than TRBA, but is still significantly faster than ABINet. Meanwhile, PARSeq-Ti achieves a much higher word accuracy vs CRNN in spite of similar parameter count and FLOPS. PARSeq-S is Pareto-optimal, while -Ti is a compelling alternative for low-resource applications.

\begin{table*}[t]
%   \scriptsize
  \fontsize{6pt}{7.2pt}\selectfont
  \centering
  \setlength\tabcolsep{1.3pt}
  \caption{Word accuracy on the six benchmark datasets (36-char). For \textit{Train data}: Synthetic datasets (\textbf{S}) - MJ \cite{Jaderberg14c} and ST \cite{Gupta16}; Benchmark datasets (\textbf{B}) - SVT, IIIT5k, IC13, and IC15; Real datasets (\textbf{R}) - COCO, RCTW17, Uber, ArT, LSVT, MLT19, ReCTS, TextOCR, and OpenVINO; "*" denotes usage of character-level labels. In our experiments, bold indicates the highest word accuracy per column. \textsuperscript{1}Used with SCATTER \cite{Litman_2020_CVPR}. \textsuperscript{2}SynthText without special characters (5.5M samples). \textsuperscript{3}LM pretrained on WikiText-103 \cite{DBLP:conf/iclr/MerityX0S17}. Combined accuracy values are available in Appendix K}
  \begin{tabular*}{\linewidth}{ c c c c c c c c c c c c }
    \toprule
    \multicolumn{3}{c}{} & \multicolumn{8}{c}{Test datasets and \# of samples} \\
    \cmidrule{4-11}
    & \multirow{2}{*}{Method} & Train & IIIT5k & SVT & \multicolumn{2}{c}{IC13} & \multicolumn{2}{c}{IC15} & SVTP & CUTE \\
    & & data & 3,000 & 647 & 857 & 1,015 & 1,811 & 2,077 & 645 & 288 \\
    \midrule
    \multirow{15}{*}{\rotatebox[origin=c]{90}{\textbf{Published Results}}} & PlugNet \cite{mou2020plugnet} & S & 94.4 & 92.3 & -- & 95.0 & -- & 82.2 & 84.3 & 85.0 \\
    & SRN \cite{yu2020towards} & S & 94.8 & 91.5 & 95.5 & -- & 82.7 & -- & 85.1 & 87.8 \\
    & RobustScanner \cite{yue2020robustscanner} & S,B & 95.4 & 89.3 & -- & 94.1 & -- & 79.2 & 82.9 & 92.4 \\
    & TextScanner \cite{wan2020textscanner} & S* & 95.7 & 92.7 & -- & 94.9 & -- & 83.5 & 84.8 & 91.6 \\
    & AutoSTR \cite{zhang2020autostr} & S & 94.7 & 90.9 & -- & 94.2 & 81.8 & -- & 81.7 & -- \\
    & RCEED \cite{cui_rceed} & S,B & 94.9 & 91.8 & -- & -- & -- & 82.2 & 83.6 & 91.7 \\
    & PREN2D \cite{Yan_2021_CVPR} & S & 95.6 & 94.0 & 96.4 & -- & 83.0 & -- & 87.6 & 91.7 \\
    & VisionLAN \cite{Wang_2021_ICCV_visionlan} & S & 95.8 & 91.7 & 95.7 & -- & 83.7 & -- & 86.0 & 88.5 \\
    & Bhunia \etal \cite{Bhunia_2021_ICCV_joint} & S & 95.2 & 92.2 & -- & 95.5 & -- & 84.0 & 85.7 & 89.7 \\
    & CVAE-Feed.\textsuperscript{1} \cite{Bhunia_2021_ICCV_towards} & S & 95.2 & -- & -- & 95.7 & -- & 84.6 & 88.9 & 89.7 \\
    % & Loginov \cite{loginov2021you} & MJ,OV & 93.5 & 94.7 & -- & 96.8 & -- & 80.2 & 89.9 & -- \\
    & STN-CSTR \cite{https://doi.org/10.48550/arxiv.2102.10884} & S & 94.2 & 92.3 & 96.3 & 94.1 & 86.1 & 82.0 & 86.2 & -- \\
    & ViTSTR-B \cite{atienza2021vitstr} & S\textsuperscript{2} & 88.4 & 87.7 & 93.2 & 92.4 & 78.5 & 72.6 & 81.8 & 81.3 \\
    & CRNN \cite{Baek_2021_CVPR} & S & 84.3 & 78.9 & -- & 88.8 & -- & 61.5 & 64.8 & 61.3 \\
    & TRBA \cite{Baek_2021_CVPR} & S & 92.1 & 88.9 & -- & 93.1 & -- & 74.7 & 79.5 & 78.2 \\
    & ABINet \cite{Fang_2021_CVPR} & S\textsuperscript{3} & 96.2 & 93.5 & 97.4 & -- & 86.0 & -- & 89.3 & 89.2 \\
    \midrule
    \multirow{12}{*}{\rotatebox[origin=c]{90}{\textbf{Experiments}}} & ViTSTR-S & S & 94.0{\tiny$\pm$0.2} & 91.7{\tiny$\pm$0.4} & 95.1{\tiny$\pm$0.7} & 94.2{\tiny$\pm$0.7} & 82.7{\tiny$\pm$0.1} & 78.7{\tiny$\pm$0.1} & 83.9{\tiny$\pm$0.6} & 88.2{\tiny$\pm$0.6} \\
    & CRNN & S & 91.2{\tiny$\pm$0.2} & 85.7{\tiny$\pm$0.7} & 92.1{\tiny$\pm$0.7} & 90.9{\tiny$\pm$0.5} & 74.4{\tiny$\pm$1.0} & 70.8{\tiny$\pm$0.9} & 73.5{\tiny$\pm$0.6} & 78.7{\tiny$\pm$0.7} \\
    & TRBA & S & 96.3{\tiny$\pm$0.2} & \textbf{92.8}{\tiny$\pm$0.9} & 96.3{\tiny$\pm$0.3} & 95.0{\tiny$\pm$0.4} & 84.3{\tiny$\pm$0.1} & 80.6{\tiny$\pm$0.2} & 86.9{\tiny$\pm$1.3} & 91.3{\tiny$\pm$1.6} \\
    & ABINet & S & 95.3{\tiny$\pm$0.2} & \textbf{93.4}{\tiny$\pm$0.2} & \textbf{97.1}{\tiny$\pm$0.4} & 95.0{\tiny$\pm$0.3} & 83.1{\tiny$\pm$0.3} & 79.1{\tiny$\pm$0.2} & 87.1{\tiny$\pm$0.6} & 89.7{\tiny$\pm$2.3} \\
    & PARSeq$_{N}$ (Ours) & S & 95.7{\tiny$\pm$0.2} & 92.6{\tiny$\pm$0.3} & 96.3{\tiny$\pm$0.4} & 95.5{\tiny$\pm$0.6} & 85.1{\tiny$\pm$0.1} & 81.4{\tiny$\pm$0.1} & 87.9{\tiny$\pm$0.9} & 91.4{\tiny$\pm$1.5} \\
    & PARSeq$_{A}$ (Ours) & S & \textbf{97.0}{\tiny$\pm$0.2} & \textbf{93.6}{\tiny$\pm$0.4} & \textbf{97.0}{\tiny$\pm$0.3} & \textbf{96.2}{\tiny$\pm$0.4} & \textbf{86.5}{\tiny$\pm$0.2} & \textbf{82.9}{\tiny$\pm$0.2} & \textbf{88.9}{\tiny$\pm$0.9} & \textbf{92.2}{\tiny$\pm$1.2} \\
    \cmidrule{2-11}
    & ViTSTR-S & R & 98.1{\tiny$\pm$0.2} & 95.8{\tiny$\pm$0.4} & 97.6{\tiny$\pm$0.3} & 97.7{\tiny$\pm$0.3} & 88.4{\tiny$\pm$0.4} & 87.1{\tiny$\pm$0.3} & 91.4{\tiny$\pm$0.2} & 96.1{\tiny$\pm$0.4} \\
    & CRNN & R & 94.6{\tiny$\pm$0.2} & 90.7{\tiny$\pm$0.4} & 94.1{\tiny$\pm$0.4} & 94.5{\tiny$\pm$0.3} & 82.0{\tiny$\pm$0.2} & 78.5{\tiny$\pm$0.2} & 80.6{\tiny$\pm$0.3} & 89.1{\tiny$\pm$0.4} \\
    & TRBA & R & 98.6{\tiny$\pm$0.1} & 97.0{\tiny$\pm$0.2} & 97.6{\tiny$\pm$0.3} & 97.6{\tiny$\pm$0.2} & 89.8{\tiny$\pm$0.4} & 88.7{\tiny$\pm$0.4} & 93.7{\tiny$\pm$0.3} & \textbf{97.7}{\tiny$\pm$0.2} \\
    & ABINet & R & 98.6{\tiny$\pm$0.2} & \textbf{97.8}{\tiny$\pm$0.3} & \textbf{98.0}{\tiny$\pm$0.4} & 97.8{\tiny$\pm$0.2} & 90.2{\tiny$\pm$0.2} & 88.5{\tiny$\pm$0.2} & 93.9{\tiny$\pm$0.8} & \textbf{97.7}{\tiny$\pm$0.7} \\
    & PARSeq$_{N}$ (Ours) & R & 98.3{\tiny$\pm$0.1} & \textbf{97.5}{\tiny$\pm$0.4} & \textbf{98.0}{\tiny$\pm$0.1} & \textbf{98.1}{\tiny$\pm$0.1} & 89.6{\tiny$\pm$0.2} & 88.4{\tiny$\pm$0.4} & 94.6{\tiny$\pm$1.0} & \textbf{97.7}{\tiny$\pm$0.9} \\
    & PARSeq$_{A}$ (Ours) & R & \textbf{99.1}{\tiny$\pm$0.1} & \textbf{97.9}{\tiny$\pm$0.2} & \textbf{98.3}{\tiny$\pm$0.2} & \textbf{98.4}{\tiny$\pm$0.2} & \textbf{90.7}{\tiny$\pm$0.3} & \textbf{89.6}{\tiny$\pm$0.3} & 
    \textbf{95.7}{\tiny$\pm$0.9} & \textbf{98.3}{\tiny$\pm$0.6} \\
    \bottomrule
  \end{tabular*}
  \label{tab:main-results}
\end{table*}

\section{Conclusion}

We adapted PLM for STR in order to learn PARSeq, a unified STR model capable of context-free and -aware decoding, and iterative refinement. PARSeq achieves SOTA results in different charset sizes and real-world datasets by jointly conditioning on both image and text representations. By unifying different decoding schemes into a single model and taking advantage of the parallel computations in Transformers, PARSeq is optimal on accuracy vs parameter count, FLOPS, and latency. Due to its extensive use of \textit{attention}, it also demonstrates robustness on vertical and rotated text common in many real-world images.

\textbf{Acknowledgments.} This work was funded in part by CHED-PCARI IIID-2016-005 (Project AIRSCAN). We are also grateful to the PCARI Prime team, led by Roel Ocampo, who ensured the uptime of our GPU servers.

\clearpage
\bibliographystyle{splncs04}
\bibliography{egbib}

% ---- arXiv version ----
\clearpage
\appendix
\section{Issues with unidirectionality of AR models in STR}

As discussed in the main text, the unidirectionality of AR models could result in spurious addition of suffixes and direction-dependent decoding. Shown in \Cref{tab:spurious-suffixes} is a sample output of a \textit{left-to-right} (LTR) AR model trained on a 36-character lowercase charset. Since the input is fairly clear and horizontal, the model was very confident in the predictions for the first 10 characters. However, since it was trained on alphanumeric characters only, it did not know how to recognize the exclamation mark. The language context \textit{swayed} the output of the model to add the \textit{-ly} suffix in order to make sense of the unrecognized character. A \textit{right-to-left} (RTL) AR model would not add the suffix due to the lack of context (since the right-most characters would have to be predicted first). This direction-dependent decoding is further illustrated in \Cref{tab:direction-dependent} where two AR models trained on opposing directions produce different outputs. In this case, the input contains ambiguity on the uppercase \textit{N} character. If read from left to right, the context of the earlier characters can be used to infer that the ambiguous character is \textit{N}. However, when read in the opposite direction, the context of \textit{OPE} is not yet available, prompting the RTL model to recognize two \textit{l}'s in place of a single \textit{N} character.

\begin{table}[htbp]
    \centering
    \setlength{\tabcolsep}{5pt}
    \caption{Example of a spurious suffix from a left-to-right AR model. \textit{GT} refers to the ground truth label, while \textit{Confidence} pertains to per-character prediction confidence}
    \label{tab:spurious-suffixes}
    \begin{tabular}{c c | c c}
        \toprule
         Input & GT & Prediction & Confidence  \\
        \midrule
         \includegraphics[width=1in]{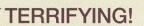} & terrifying & terrifying\textcolor{red}{ly} & [1.00, \dots, 1.00, 0.97, 0.72] \\
        \bottomrule
    \end{tabular}
\end{table}

\begin{table}[htbp]
    \centering
    \setlength{\tabcolsep}{5pt}
    \caption{Example of direction-dependent decoding with two AR models}
    \label{tab:direction-dependent}
    \begin{tabular}{c c | c c c}
        \toprule
        Input & GT & Direction & Prediction & Confidence \\
        \midrule
        \multirow{2}{*}{\includegraphics[width=0.8in]{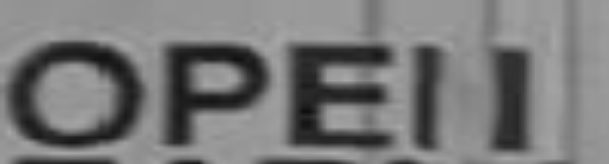}} & \multirow{2}{*}{open} & LTR & open & [1.00, 1.00, 1.00, 0.66] \\
        & & RTL & ope\textcolor{red}{ll} & [1.00, 1.00, 0.52, 0.57, 0.94] \\
        \bottomrule
    \end{tabular}
\end{table}

\section{Inefficiency of External Language Models in STR}

As mentioned in the main text, ensemble methods such as ABINet \cite{Fang_2021_CVPR} and SRN \cite{yu2020towards} utilize a standalone or external Language Model (LM). In \Cref{tab:external-lm}, we show the cost measurements of \texttt{fvcore} on the full ABINet model for a single input, as well as the measurement breakdown for its component models. We can see that while the LM accounts for around 34.48\% of the parameter count, it only uses 13.65\% of the overall FLOPS and 15.78\% of the overall activations (a measure shown to be correlated with model runtime \cite{dollar2021fast,xiao2021early}). When evaluated in spelling correction on the 36-character set, the LM achieves a top-5 word accuracy of only 41.9\% \cite{Fang_2021_CVPR}. With the ground truth label itself as input (\Cref{tab:abinet-lm-acc}), the same model gets a top-1 word accuracy of only 50.44\% (36-char). This means that even if the Vision Model (VM) is perfect (always predicting the correct label), the LM will produce a wrong output 50\% of the time. In summary, the external LM's dedicated compute cost, underutilization relative to its parameter and memory requirements, and dismal word accuracy show the inefficiency of this approach. For STR, an internal LM might be more appropriate since the primary input signal is the image, not the language context.

\begin{table}[htbp]
    \setlength{\tabcolsep}{6pt}
    \centering
    \fontsize{8}{9.6}\selectfont
    \caption{Commonly used cost indicators as measured by \texttt{fvcore} for ABINet. \textit{Full Model} pertains to the overall measurements}
    \label{tab:external-lm}
    \begin{tabular}{l|r r r}
        \toprule
        \textbf{Module} & \# of Parameters (M) & FLOPS (G) & \# of Activations (M) \\
        \midrule
        Full Model & 36.858 \textit{(100.00\%)} & 7.289 \textit{(100.00\%)} & 10.785 \textit{(100.00\%)} \\
        \midrule
        - Vision & 23.577 \textit{(63.97\%)} & 6.249 \textit{(85.73\%)} & 9.036 \textit{(83.78\%)} \\
        - Language & 12.707 \textit{(34.48\%)} & 0.995 \textit{(13.65\%)} & 1.702 \textit{(15.78\%)} \\
        - Alignment & 0.574 \textit{(1.55\%)} & 0.045 \textit{(0.62\%)} & 0.047 \textit{(0.44\%)} \\
        \bottomrule
    \end{tabular}
\end{table}

\begin{table}[htbp]
    \centering
    \fontsize{8}{9.6}\selectfont
    \caption{Performance of ABINet's LM when the ground truth label itself is used as the input}
    \begin{tabular}{c|r|r|r}
        \toprule
        Dataset & \# of samples & Word acc. (\%) & 1 - NED \\
        \midrule
        IIIT5k & 3,000 & 47.33 & 69.50 \\
        SVT & 647 & 65.38 & 83.48 \\
        IC13 & 1,015 & 62.07 & 78.77 \\
        IC15 & 2,077 & 40.49 & 67.72 \\
        SVTP & 645 & 65.27 & 83.08 \\
        CUTE80 & 288 & 46.88 & 68.65 \\
        \midrule
        \textbf{Combined} & \textbf{7,672} & \textbf{50.44} & \textbf{72.54} \\
        \bottomrule
    \end{tabular}
    \label{tab:abinet-lm-acc}
\end{table}

\section{Multi-head Attention}

The attention mechanism is central to the operation of Transformers \cite{vaswani2017attention}. In scaled dot-product attention, the similarity scores between two $d_k$-dimensional vectors $\mathbf{q}$ (query) and $\mathbf{k}$ (key), computed using their dot-product, are used to transform a $d_v$-dimensional vector $\mathbf{v}$ (value). Formally, scaled dot-product attention is defined as:

\begin{equation}
  Attn(\mathbf{q}, \mathbf{k}, \mathbf{v}) = softmax \left( \frac{\mathbf{q}\mathbf{k}^T}{\sqrt{d_k}} \right) \mathbf{v}
  \label{eq:sdpattn}
\end{equation}It accepts an optional \textit{attention mask} that limits which \textit{keys} the \textit{queries} could attend to. In a Transformer with token dimensionality of $d_{model}$, $d_k = d_v = d_{model}$.

Multi-head Attention (MHA) is the extension of scaled dot-product attention to multiple representation subspaces or \textit{heads}. To keep the computational cost of MHA practically constant regardless of the number of heads, the dimensionality of the vectors are reduced to $d_{head} = d_{model} / h$, where $h$ is the number of heads. A \textit{head} corresponds to an invocation of \Cref{eq:sdpattn} on projected versions of $\mathbf{q}$, $\mathbf{k}$, and $\mathbf{v}$ using parameter matrices $\mathbf{W}^q \in \mathbb{R}^{d_{model}\times{d_{head}}}$, $\mathbf{W}^k \in \mathbb{R}^{d_{model}\times{d_{head}}}$, and $\mathbf{W}^v \in \mathbb{R}^{d_{model}\times{d_{head}}}$, respectively, as shown in \Cref{eq:attn_head}. The final output is obtained in \Cref{eq:mha} by concatenating the heads and multiplying by the output projection matrix $\mathbf{W}^o \in \mathbb{R}^{{d_{model}}\times{d_{model}}}$.

\begin{equation}
    head_i = Attn(\mathbf{q}\mathbf{W}^q_i, \mathbf{k}\mathbf{W}^k_i, \mathbf{v}\mathbf{W}^v_i)
    \label{eq:attn_head}
\end{equation}

\begin{equation}
    MHA(\mathbf{q}, \mathbf{k}, \mathbf{v}) = Concat(head_1, ..., head_h)\mathbf{W}^o
  \label{eq:mha}
\end{equation}

\section{Model Architecture}
\label{ch:arch-appendix}

PARSeq uses an encoder which largely follows the original ViT \cite{dosovitskiy2020image}, and a pre-$LayerNorm$ \cite{baevski2018adaptive,wang2019learning} decoder with more heads. The architectures are practically unchanged but are reproduced here for the convenience of the reader.

\subsection{ViT Encoder}

\begin{figure}[ht]
  \centering
  \includegraphics[height=0.58\linewidth]{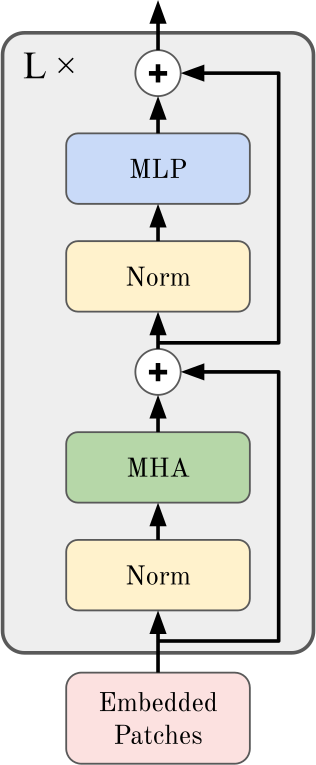}
   \caption[Illustration of a ViT layer.]{Illustration of a ViT layer from Dosovitskiy \etal \cite{dosovitskiy2020image}. \textit{Norm} pertains to $LayerNorm$.}
   \label{fig:vit-arch}
\end{figure}

The encoder is composed of 12 layers. All layers share the same architecture shown in \Cref{fig:vit-arch}. The output of the last encoder layer goes through a final $LayerNorm$.

\subsection{Visio-lingual Decoder}

\begin{figure}[ht]
  \centering  \includegraphics[width=0.98\linewidth]{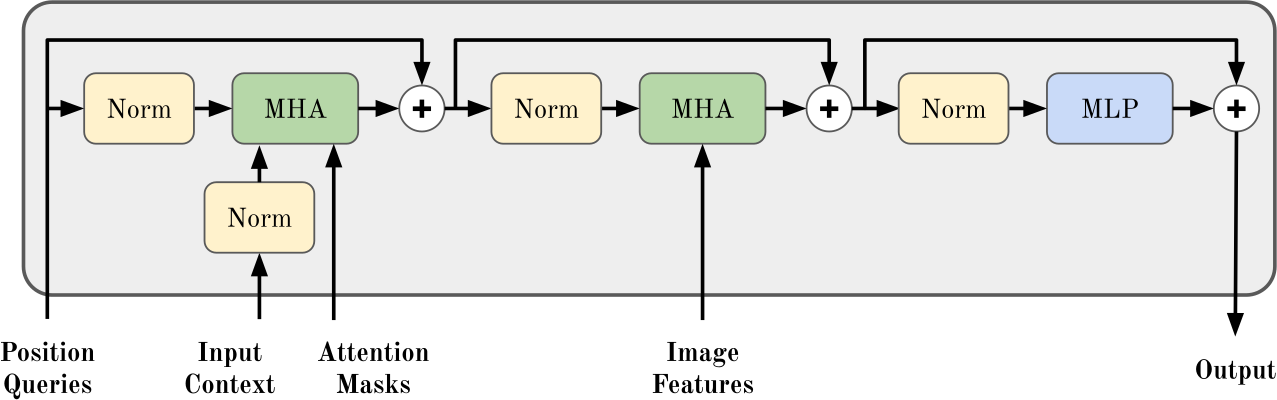}
   \caption{Visio-lingual decoder architecture with $LayerNorm$ layers shown.}
   \label{fig:decoder-arch-detailed}
\end{figure}

The decoder (\Cref{fig:decoder-arch-detailed}) consists of only a single layer. The immediate outputs of all $MHA$ and $MLP$ layers go through $Dropout$ ($p = 0.1$, not shown). \textit{Image Features} are already $LayerNorm$'d by the encoder (hence no $LayerNorm$ prior to input).

\subsection{Architecture Configuration}
\label{sec:arch-config}

The main results are obtained from the base model, PARSeq-S, which has a similar configuration to DeiT-S \cite{touvron2021training} but uses an image size of 128$\times$32 and a patch size of 8$\times$4 (a change also adapted in our reproduction of ViTSTR-S). Based on our experiments, scaling up the model only marginally improves word accuracy on the benchmark. We instead explore scaling down the model to make it more suitable for edge devices. PARSeq-Ti, which uses a configuration similar to DeiT-Ti \cite{touvron2021training}, is more similar to CRNN \cite{shi2016end} in terms of parameter count and FLOPS. The detailed configuration parameters are shown in \Cref{tab:arch-config}.

\begin{table}[ht]
  \centering
  \setlength{\tabcolsep}{8pt}
  \caption[Configurations for PARSeq-S and PARSeq-Ti model variants.]{Configurations for the base (PARSeq-S) and smaller (PARSeq-Ti) model variants. $d_{model}$ refers to the \textit{dimensionality} of the model which dictates the dimensions of the vectors and feature maps. $h$ refers to the number of \textit{attention heads} used in MHA layers. $d_{MLP}$ refers to the dimension of the intermediate features within the MLP layer. \textit{depth} refers to the number of encoder or decoder layers used}
  \begin{tabular}{c|c|c c c|c c c}
    \toprule
    & & \multicolumn{3}{c}{\textit{encoder}} & \multicolumn{3}{c}{\textit{decoder}} \\
    \textbf{Variants} & $d_{model}$ & $h$ & $d_{MLP}$ &
      depth & $h$ & $d_{MLP}$ & depth \\
    \midrule
      PARSeq-Ti & 192 & 3 & 768 & 12 & 6 & 768 & 1 \\
      PARSeq-S & 384 & 6 & 1536 & 12 & 12 & 1536 & 1 \\
    \bottomrule
  \end{tabular}
  \label{tab:arch-config}
\end{table}

\section{Permutation Language Modeling}
\label{sec:plm-appendix}

In this section, we provide additional details about the adaptation of PLM for use in PARSeq. We give a concrete illustration of masked multi-head attention first. Next, the intuition behind the usage of permutation pairs is discussed. Lastly, implementation details and considerations about the training procedure are discussed.

\subsection{Illustration of attention masking}

\begin{figure}[htbp]
  \centering
    \subfloat[MHA for output token $y_1$]{%
         \centering
         \includegraphics[width=0.45\textwidth]{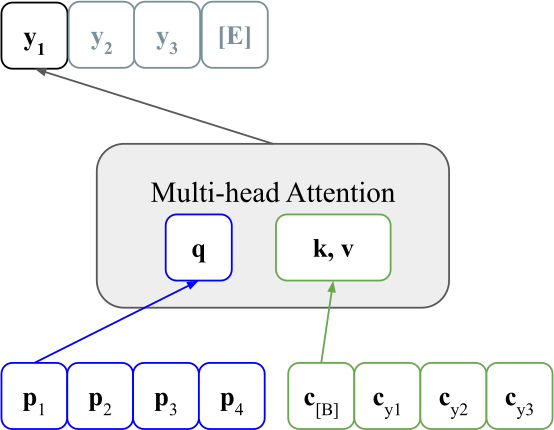}
         \label{fig:attn-connectivity:y1}
    }
    \hfill
    \subfloat[MHA for output token $y_2$]{%
         \centering
         \includegraphics[width=0.45\textwidth]{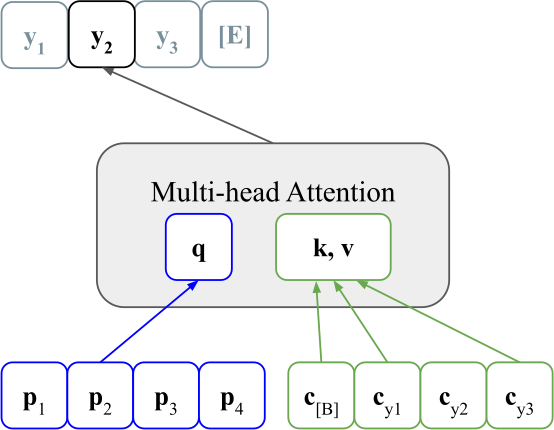}
         \label{fig:attn-connectivity:y2}
     }
     \hfill
     \subfloat[MHA for output token $y_3$]{%
         \centering
         \includegraphics[width=0.45\textwidth]{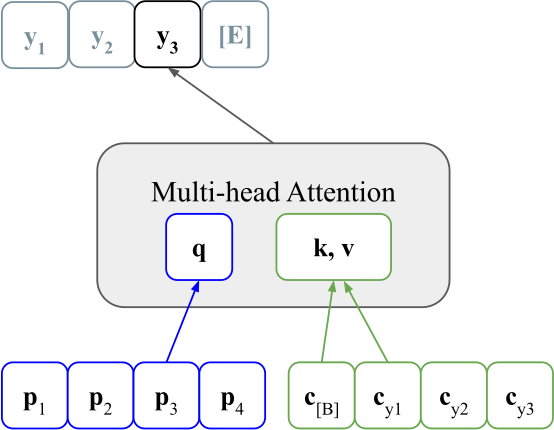}
         \label{fig:attn-connectivity:y3}
     }
    \hfill
    \subfloat[MHA for output token \texttt{[E]}]{%
         \centering
         \includegraphics[width=0.45\textwidth]{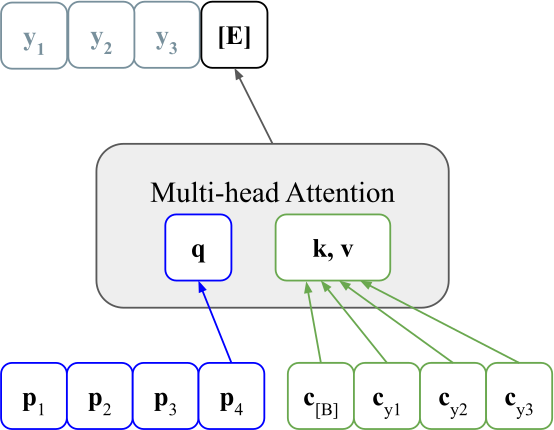}
         \label{fig:attn-connectivity:eos}
     }
     
     \caption{Masked MHA for a three-element sequence $\mathbf{y} = [y_1, y_2, y_3]$ given the factorization order $[1, 3, 2]$. $\mathbf{c}$ are context embeddings with position information}
    \label{fig:attn-connectivity}
\end{figure}

As discussed in the main text, Transformers process all tokens in parallel. In order to enforce the \textit{AR} constraint which limits the conditional dependencies for each token, attention masking is used. \Cref{fig:attn-connectivity} shows a concrete example of masked multi-head attention for a sequence $\mathbf{y}$. The \textit{position} tokens always serve as the \textit{query} vectors, while the \textit{context} tokens (context \textit{embeddings} with position information) serve as the \textit{key} and \textit{value} vectors. Note that the sequence order is \textit{fixed}, and that only the AR factorization order (specified by the attention mask) is permuted.

\subsection{Permutation Sampling}
\label{sec:perm-sampling}

As discussed in the main text, we sample permutations in a specific way. We use pairs of permutations, and the \textit{left-to-right} permutation is always used. Thus, we only sample $K / 2 - 1$ permutations every training step. To illustrate the intuition behind the usage of \textit{flipped} permutation pairs, we give the following example. Given a three-element text label $\mathbf{y} = [y_1, y_2, y_3]$ and $K = 4$ permutations: $[1, 2, 3]$, $[3, 2, 1]$, $[1, 3, 2]$, and $[2, 3, 1]$. The first two permutations are the \textit{left-to-right} and \textit{right-to-left} orderings, respectively. Both are always used as long as $K > 1$. The corresponding factorizations of the joint probability per pair are as follows:

\begin{align*}
    p(\mathbf{y})_{[1, 2, 3]} &= p(y_1) p(y_2 | y_1) p(y_3 | y_1, y_2) \\
    p(\mathbf{y})_{[3, 2, 1]} &= p(y_3) p(y_2 | y_3) p(y_1 | y_2, y_3)
\end{align*}
    
\begin{align*}
    p(\mathbf{y})_{[1, 3, 2]} &= p(y_1) p(y_3 | y_1) p(y_2 | y_1, y_3) \\
    p(\mathbf{y})_{[2, 3, 1]} &= p(y_2) p(y_3 | y_2) p(y_1 | y_2, y_3)
\end{align*}

For each permutation pair, if we group the probabilities per element, we get \Cref{tab:grouped-probs}. Notice that the probabilities of each element for every permutation pair consists of disjoint sets of conditioning variables. For example, the probabilities of element $y_1$ for $[1,2,3]$  (\textit{left-to-right} permutation) and $[3,2,1]$ (\textit{right-to-left} permutation) are $p(y_1)$ and $p(y_1 | y_2, y_3)$, respectively. The first term is the prior probability of $y_1$. It is not conditioned on any other element of the text label, unlike the second term which is conditioned on all other elements, $y_2$ and $y_3$. Similarly for $y_2$, the first term is conditioned only on $y_1$ while the second term is conditioned only on $y_3$. In our experiments, we find that using flipped permutation pairs results in more stable training dynamics where the loss is smoother and less erratic.

\begin{table}[ht]
  \centering
  \setlength{\tabcolsep}{6pt}
  \caption{Probability terms grouped by permutation pairs.}
  \begin{tabular}{ c | l l l }
    \toprule
    Perm. & $y_1$ & $y_2$ & $y_3$ \\
    \midrule
    $[1,2,3]$ & $p(y_1)$ & $p(y_2 | y_1)$ & $p(y_3 | y_1, y_2)$ \\
    $[3,2,1]$ & $p(y_1 | y_2, y_3)$ & $p(y_2 | y_3)$ & $p(y_3)$ \\
    \midrule
    $[1,3,2]$ & $p(y_1)$ & $p(y_2 | y_1, y_3)$ & $p(y_3 | y_1)$ \\
    $[2,3,1]$ & $p(y_1 | y_2, y_3)$ & $p(y_2)$ & $p(y_3 | y_2)$ \\
    \bottomrule
  \end{tabular}
  \label{tab:grouped-probs}
\end{table}

% \subsection{Generation of Attention Masks from Permutations}

% Attention masking is used to enforce the autoregressive property of the model by limiting the flow of information during training. A sampled permutation determines the factorization order

\subsection{Special handling of end-of-sequence \texttt{[E]} token}

Although the \texttt{[E]} token is part of the sequence, it is handled in a specific way in order to make training simpler. First, no character $c \in C$, where $C$ is the training charset, is conditioned on \texttt{[E]}. Intuitively, it means that \texttt{[E]} marks the end of the sequence (hence its name) since no more characters are expected after it is produced by the model. More formally, it means that $p(c | \texttt{[E]}) = 0$. This is achieved by masking the positions of \texttt{[E]} in the input context. Second, we train \texttt{[E]} on only two permutations, \textit{left-to-right} and \textit{right-to-left}. The \textit{left-to-right} lookahead mask provides the longest context to \texttt{[E]} (conditioned on all other characters in the sequence), while the \textit{right-to-left} mask provides no context, which is necessary for NAR decoding. We could also train \texttt{[E]} on different subsets of the input context, but doing so needlessly complicates the training procedure without offering any advantages.

\subsection{Considerations for batched training}

Text labels of varying lengths can be included in a mini-batch. However, the sampled permutations for the mini-batch are always based on the longest sequence. Hence, it is possible that after accounting for padding, multiple permutations would become equivalent. To see why this is the case, consider a mini-batch containing two samples: the first label has a single character, while the second label has four characters. The first label has a sequence length of one and total number of permutations also equal to one. On the other hand, the second label has a sequence length of four which corresponds to 24 total permutations. If we use $K = 6$ permutations, then it means that the permutations for the first label would be oversampled since there is only one valid permutation for $T = 1$. We find that this oversampling actually helps training. We experimented with a modified training procedure wherein sequences with $T < 4$ are grouped together (\ie 1-, 2-, and 3-character sequences are grouped separately). This training procedure results in increased training time due to the mini-batch being split further into smaller batches, but it does not improve accuracy nor hasten convergence. Thus, we stick with the simpler batched training procedure.

% -----------------------------------------------------
\section{Dataset Matters}

\subsection{Open Images Datasets}
\label{ch:open-images}

\begin{figure}[htbp]
  \centering
  \includegraphics[width=0.9\linewidth]{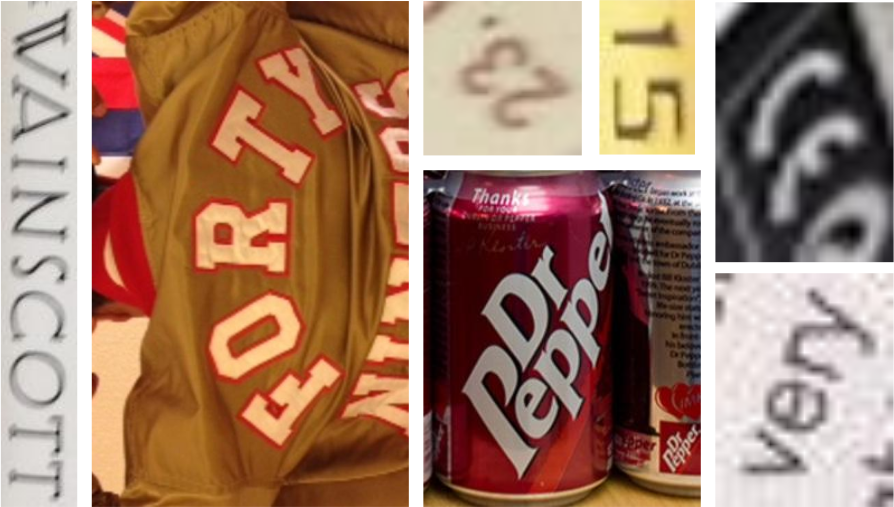}
   \caption{Cropped word boxes from Open Images.}
   \label{fig:open-images-cropped-words}
\end{figure}

\begin{figure*}[htbp]
  \centering
  \includegraphics[width=\linewidth]{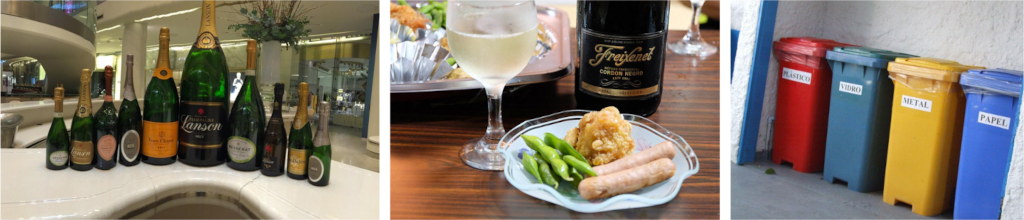}
   \caption{Examples of source scene images common to TextOCR and OpenVINO.}
   \label{fig:textocr-openvino-overlap}
\end{figure*}

TextOCR and OpenVINO are datasets both derived from Open Images---a large dataset with very diverse images often containing complex scenes with several objects (8.4 per image on average). Open Images is not specifically collected for STR. Thus, it contains text of varying resolutions, orientations, and quality, as shown in cropped word boxes in \Cref{fig:open-images-cropped-words}. TextOCR and OpenVINO significantly overlap in terms of source scene images, as shown in \Cref{tab:textocr-openvino-overlap}. Samples of source scene images common to both are shown in \Cref{fig:textocr-openvino-overlap}. Only the 
\textit{validation} set of OpenVINO and the \textit{test} set of TextOCR do not overlap any other image set. The labels of TextOCR's \textit{test} set are kept private.

\begin{table}[htbp]
  \centering
  \setlength{\tabcolsep}{10pt}
  \caption{Overlap between TextOCR and OpenVINO in terms of the number of common source scene images.}
  \begin{tabular}{ c c | r r r }
    \toprule
    & & \multicolumn{3}{c}{\textbf{TextOCR}} \\
    & & train & val & test \\
    \midrule
    \multirow{5}{*}{\rotatebox[origin=c]{90}{\textbf{OpenVINO}}} & train\_1 & 1,612 & 225 & 0 \\
    & train\_2 & 1,444 & 230 & 0 \\
    & train\_5 & 1,302 & 184 & 0 \\
    & train\_f & 1,068 & 157 & 0 \\
    & validation & 0 & 0 & 0 \\
    \bottomrule
  \end{tabular}
  \label{tab:textocr-openvino-overlap}
\end{table}

\subsection{Data preparation for LMDB storage}

We use the archives released by Baek \etal \cite{Baek_2021_CVPR} for RCTW17, Uber-Text, ArT, LSVT, MLT19, and ReCTS. Thus, we only preprocess data for the remaining datasets.

For COCO-Text, we use the v1.4 \textit{test} annotations released as part of the ICDAR 2017 challenge. For \textit{train} and \textit{val}, we use the latest (v2.0) annotations. We preprocess TextOCR, OpenVINO, and COCO-Text with minimal filtering and modifications, in contrast to the usual practice of removing non-horizontal text and special characters. We only filter illegible and non-machine printed text. The only modification we perform is the removal of whitespace on either side of the label, or duplicate whitespace between non-whitespace characters.

For IC13 and IC15, we use the original data from the ICDAR competition website and perform no modifications to the data. We emulate the previous filtering methods \cite{wang2011end,cheng2017focusing} to create the subsets used for evaluation.

Long and Yao \cite{long2020unreal} have reannotated IIIT5k, CUTE, SVT, and SVTP because the original annotations are case-insensitive and lack punctuation marks. However, both the reannotations and the originals contain some errors. Hence, we review inconsistencies between the two versions and manually reconcile them to correct the errors.

\Cref{tab:dataset} provides a detailed summary of how each dataset was used.

\begin{table}[htbp]
  \centering
  \fontsize{6.5}{7.8}\selectfont
  \caption{Summary of dataset usage after on-the-fly filtering for the 94-character set. Numbers indicate how many samples were used from each dataset. \textsuperscript{t} and \textsuperscript{v} refer to splits that were repurposed as training and validation data, respectively. \textsuperscript{*} indicates private ground truth labels. -- indicates that the dataset does not have a particular split. IC13 and IC15 have two \textit{versions} of their respective \textit{test} splits commonly used in the literature.}
  \setlength{\tabcolsep}{10pt}
  \begin{tabular}{ c | r r r }
    \toprule
    Dataset & \textit{train} & \textit{val} & \textit{test} \\
    \midrule
    MJSynth & 7,224,586 & 802,731\textsuperscript{t} & 891,924\textsuperscript{t} \\
    SynthText & 6,975,301 & -- & -- \\
    \midrule
    LSVT & 41,439 & -- & -- \\
    MLT19 & 56,727 & -- & -- \\
    RCTW17 & 10,284 & -- & -- \\
    ReCTS & 21,589 & -- & 2,467\textsuperscript{t} \\
    TextOCR & 710,994 & 107,093\textsuperscript{t} & 0\textsuperscript{*} \\
    OpenVINO & 1,912,784 & 158,757\textsuperscript{t} & -- \\
    \midrule
    ArT & 32,028 & -- & 35,149 \\
    COCO & 59,733 & 13,394\textsuperscript{t} & 9,825 \\
    Uber & 91,732 & 36,188\textsuperscript{t} & 80,587 \\
    \midrule
    IIIT5k & 2,000\textsuperscript{v} & -- & 3,000 \\
    SVT & 257\textsuperscript{v} & -- & 647 \\
    IC13 & 848\textsuperscript{v} & -- & 857 / 1,015 \\
    IC15 & 4,468\textsuperscript{v} & -- & 1,811 / 2,077 \\
    SVTP & -- & -- & 645 \\
    CUTE & -- & -- & 288 \\
    \bottomrule
  \end{tabular}
  \label{tab:dataset}
\end{table}

% ----------------------------------------
\section{Training Details}

In the main text, the 169,680 training iterations (batch size of 384) is equivalent to 20 epochs on the combined real training dataset (3,257,585 samples). The same exact number of training iterations is used when training on synthetic data (MJ+ST, 15.89M samples), resulting in just over 4 epochs of training. As shown in \Cref{tab:training-sched}, this training schedule is more than twice as long as Baek \etal \cite{Baek_2021_CVPR} but is still much shorter than ABINet's original training schedule of 8 epochs VM pretraining on MJ+ST, 80 epochs LM pretraining on WikiText-103, and 10 epochs full model training on MJ+ST. This explains why our reproduction of CRNN and TRBA obtain higher accuracy than the originals, and why ABINet gets a slightly lower (1.4\%) accuracy compared to the original results.

\begin{table}[htbp]
    \centering
    \setlength{\tabcolsep}{5pt}
    \caption{Training schedule comparison vs reproduced methods. Sorted from shortest to longest schedule based on the \textit{sample count} (essentially \textit{batch size} $\times$ \textit{number of iterations})}
    \label{tab:training-sched}
    \begin{tabular}{c r r | r}
        \toprule
        \textbf{Method} & \textbf{Batch size} & \textbf{\# of iterations} & \textbf{Sample count} (M) \\
        \midrule
        CRNN and TRBA \cite{Baek_2021_CVPR} & 128 & 200,000 & 25.6 \\
        ViTSTR \cite{atienza2021vitstr} & 192 & 300,000 & 57.6 \\
        \textit{Ours} & 384 & 169,680 & 65.2 \\
        ABINet (VM + full) \cite{Fang_2021_CVPR} & 384 &  745,074 & 286.1 \\
        ABINet (LM) \cite{Fang_2021_CVPR} & 4,096 & 1,688,720 & 6,917.0 \\
        \bottomrule
    \end{tabular}
\end{table}

\subsection{Label preprocessing}
Preprocessing and filtering are done as follows. Whitespace characters are removed from the labels. Unicode characters are normalized using the NFKD normalization form and then converted to ASCII. Next, labels longer than $T$ characters are filtered. Case-sensitivity is inferred from the charset. If all letters in the charset are lowercase, the label is transformed to its lowercase version. If the charset consists of purely uppercase letters, the label is converted to its uppercase version. If the charset is mixed-case, no case conversion is done. Lastly, all characters not specified in the charset are removed from the labels.

\subsection{Learning Rate Optimization}
\label{ch:learning-rates}

\begin{table}[htbp]
  \centering
  \fontsize{7}{8.4}\selectfont
  \setlength{\tabcolsep}{5pt}
  \caption[Learning rates used for training.]{Learning rates used for training. The \textit{Base LR} is the raw value set in the configuration, while the \textit{Effective LR} is the actual value used for training. During pretraining, \textit{ABINet (LM)} is used for ABINet's language model}
  \begin{tabular}{ c | r r }
    \toprule
    Model & Base LR & Effective LR \\
    \midrule
    CRNN & $5.10\times10^{-4}$ & $1.08\times10^{-3}$ \\
    ViTSTR-S & $8.90\times10^{-4}$ & $1.89\times10^{-3}$ \\
    TRBC & $1.00\times10^{-4}$ & $2.12\times10^{-4}$ \\
    TRBA & $6.90\times10^{-4}$ & $1.46\times10^{-3}$ \\
    ABINet (LM) & $3.00\times10^{-4}$ & $6.36\times10^{-4}$ \\
    ABINet & $3.40\times10^{-4}$ & $7.21\times10^{-4}$ \\
    PARSeq & $7.00\times10^{-4}$ & $1.48\times10^{-3}$ \\
    \bottomrule
  \end{tabular}
  \label{tab:learning-rates}
\end{table}

For fair comparison during evaluation, all training hyperparameters---except the learning rate---are kept constant across models. The learning rate is varied because different architectures and model sizes train differently \cite{li2020train}. Ray Tune \cite{liaw2018tune} was used to automatically search for the optimum maximum learning rate given the fixed training schedule. Specifically, we used a combination of Median Stopping \cite{Vizier} and Bayesian Optimization \cite{balandat2020botorch} to efficiently narrow down the configuration space. Finally, a grid search over the narrowed down learning rate range was performed with models trained to completion. The final learning rates used for training are shown in \Cref{tab:learning-rates}. The base learning rates are scaled using two multipliers: the DDP factor ($\sqrt{nGPU}$) and the batch size linear scaling rule ($bsize / 256$) \cite{goyal2017accurate}, where $nGPU$ refers to the number of GPUs used (\ie $nGPU = 2$ for a dual-GPU setup) and $bsize$ refers to the batch size (\ie $bsize = 384$).

%-----------------------------------------
\section{Accuracy of decoding schemes vs latency}
\label{sec:acc-vs-latency}

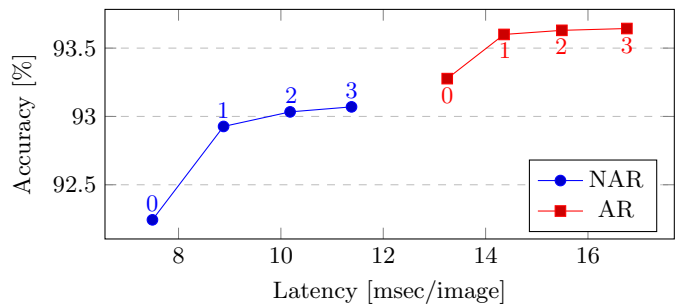
\begin{figure}[htbp]
  \centering

    \begin{tikzpicture}
    \begin{axis}[
        ylabel={Accuracy [\%]},
        xlabel={Latency [msec/image]},
        legend pos=south east,
        ymajorgrids=true,
        grid style=dashed,
        cycle list name=color,
        width=0.75\linewidth,
        height=0.38\linewidth
    ]
    
    \addplot
        coordinates {
        (7.49,92.2433) (8.88,92.9267) (10.18,93.0333) (11.38,93.0700)
        } node[above, pos = 0] {0}
          node[above, pos = 0.38] {1}
          node[above, pos = 0.71] {2}
          node[above, pos = 1] {3}
          ;
        
    \addplot
        coordinates {
        (13.25,93.2767) (14.36,93.6000) (15.49,93.6300) (16.76,93.6433)
        } node[below, pos = 0] {0}
          node[below, pos = 0.33] {1}
          node[below, pos = 0.64] {2}
          node[below, pos = 1] {3}
          ;

    \legend{NAR,AR}

    \end{axis}
    \end{tikzpicture}

  \caption[PARSeq word accuracy and single-image latency for each decoding scheme.]{PARSeq word accuracy and single-image latency for each decoding scheme. The number of refinement iterations used is indicated for each point.}
  \label{fig:ablation-decoding}
\end{figure}

\Cref{fig:ablation-decoding} shows how the word accuracy and latency evolve as functions of the number of refinement iterations. For AR decoding, refinement iterations after the first provide negligible increase in accuracy. For NAR decoding, the accuracy increase becomes insignificant after the second iteration. Hence, we use one and two refinement iterations for the AR and NAR decoding schemes, respectively.

\section{Detailed Latency Measurements}
\label{ch:latency-measurements}

\begin{figure}[htbp]
  \centering

    \begin{tikzpicture}
    \begin{axis}[
        xlabel={Output label length},
        ylabel={Latency [msec/image]},
        ymin=0, ymax=38,
        xmin=1, xmax=25,
        xtick={1, 5, 9, 13, 17, 21, 25},
        ytick={0, 5, 10, 15, 20, 25, 30, 35},
        % legend pos=south east,
        legend pos=outer north east,
        % ymajorgrids=true,
        grid style=dashed,
        cycle list name=color,
        width=0.8\linewidth,
        height=0.46\linewidth
    ]
    
    \addplot
        coordinates {
        (1,3.29) (5,3.12) (9,3.12) (13,3.15) (17,3.11) (21,3.12) (25,3.12)
        };
        
    \addplot
        coordinates {
        (1,5.95) (5,5.96) (9,5.94) (13,5.97) (17,5.99) (21,5.96) (25,5.98)
        };
    
    \addplot
        coordinates {
        (1,10.37) (5,11.77) (9,13.25) (13,14.69) (17,16.49) (21,17.80) (25,19.09)
        };
        
    \addplot
        coordinates {
        (1,24.79) (5,24.58) (9,24.44) (13,24.73) (17,24.38) (21,24.36) (25,24.70)
        };
        
    \addplot
        coordinates {
        (1,9.93) (5,10.06) (9,9.99) (13,10.06) (17,10.41) (21,10.41) (25,10.42)
        };
        
    \addplot
        coordinates {
        (1,9.71) (5,13.99) (9,18.19) (13,22.37) (17,27.07) (21,31.48) (25,36.22)
        };
        
    \addplot +[mark=none] coordinates {(5.5, 0) (5.51, 40)} node[right, pos = 0.85] {Mean word length};
    
    \legend{CRNN,ViTSTR-S,TRBA,ABINet,PARSeq$_{N}$,PARSeq$_{A}$}

    \end{axis}
    \end{tikzpicture}

  \caption[Model latency vs output label length.]{Model latency vs output label length as measured by PyTorch's benchmark timer on an NVIDIA Tesla A100 GPU. Each point corresponds to a mean of five runs of \texttt{Timer.blocked\_autorange()}. Lower is better. \textit{Mean word length} is the average length of the labels from all test datasets.}
  \label{fig:runtime}
\end{figure}
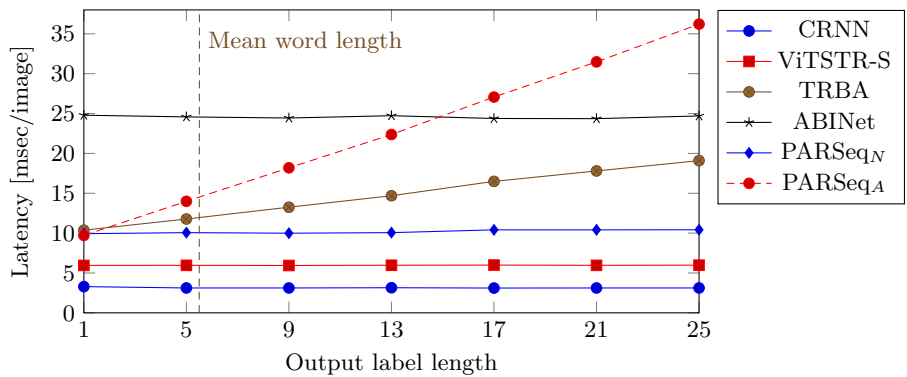

We measure model latency in an isolated manner. By doing so, we can reliably factor out the effects of data loading, storage, and CPU latency. We use the built-in benchmarking tool of PyTorch and measure latency for different label lengths, as shown in \Cref{fig:runtime}. As expected, NAR methods including PARSeq$_N$ exhibit near-constant latency regardless of output label length. Meanwhile, the latency of AR methods increases linearly as a function of the output label length. The latency increase of PARSeq$_A$ is steeper than TRBA. However, since the average length of words in the test datasets is quite short at 5.4, the actual difference in mean latency between TRBA and PARSeq$_A$ is only about 2.3 msec.

\section{Experiments on arbitrarily-oriented text}

In STR, the focus has mainly been on horizontal text, with a few explicitly tackling arbitrarily-oriented text \cite{phan2013recognizing,cheng2018aon,munjal2021stride,Yan_2021_CVPR,yan2021mean}. In our experiments, we observe that existing attention-based models are capable of recognizing text in arbitrary orientation, as shown in \Cref{tab:rotation-results}. Only CRNN, a CTC-based model, exhibits dismal orientation robustness. We conjecture that the direct correspondence of visual feature positions to textual feature positions in CTC-based models causes this poor performance. On the other hand, attention-based models compute feature similarity scores on-the-fly, resulting in a more dynamic alignment between visual and textual features.

\begin{table}[htbp]
  \centering
  \scriptsize
  \setlength{\tabcolsep}{5pt}
  \caption{Orientation robustness benchmark. Word accuracy (94-char) on rotated versions of the six benchmark datasets. \textit{\%dec} refers to the percentage decrease of the \textit{Mean} accuracy \wrt the 0\textdegree{} accuracy.}
  \begin{tabular}{ c | c | r r r | r r }
    \toprule
    Method & 0\textdegree$\uparrow$ & 90\textdegree$\uparrow$ & 180\textdegree$\uparrow$ & 270\textdegree$\uparrow$ & Mean$\uparrow$ & \%dec$\downarrow$ \\
    \midrule
    CRNN  & 85.8 & 11.8$\pm$0.1 & 6.4$\pm$0.5 & 10.7$\pm$0.2 & 9.6 & 88.8 \\
    ViTSTR-S & 91.8 & \textbf{87.9}$\pm$0.2 & 78.9$\pm$0.3 & 80.6$\pm$0.9 & 82.4 & 10.2 \\
    TRBA & 92.5 & 84.6$\pm$0.1 & 83.5$\pm$0.2 & 78.6$\pm$0.3 & 82.3 & 11.0 \\
    ABINet & 92.4 & 66.0$\pm$1.5 & 77.1$\pm$0.9 & 65.5$\pm$1.8 & 69.6 & 24.7 \\
    \midrule
    PARSeq$_{N}$ & 92.7 & 86.7$\pm$0.3 & 83.2$\pm$0.9 & 81.1$\pm$0.3 & 83.7 & 9.7 \\
    PARSeq$_{A}$ & \textbf{93.3} & \textbf{88.0}$\pm$0.1 & \textbf{86.6}$\pm$0.3 & \textbf{84.1}$\pm$0.1 & \textbf{86.2} & \textbf{7.6} \\
    \bottomrule
  \end{tabular}
  \label{tab:rotation-results}
\end{table}

We hypothesize that regardless of architecture and training procedure, the attention mechanism makes STR models generally robust against orientation variations. To test our hypothesis, we created a pose-corrected version of TextOCR which contains text in canonical orientation (practically horizontal), as opposed to the original version which contains text in arbitrary orientation. One possible contributor to the orientation robustness of TRBA is its image rectification module \cite{shi2016robust}. To test if this is the case, we also train TRBC, the CTC-based version of TRBA. We train all models on either TextOCR variants exclusively and show the results in \Cref{tab:horizontal-vs-arbitrary}.

\begin{table}[htbp]
  \centering
  \scriptsize
  \setlength{\tabcolsep}{10pt}
  \caption{Effect of training on horizontally-oriented (H) vs arbitrarily-oriented (A) variants of TextOCR. 0\textdegree{} pertains to model accuracy (94-char) on non-rotated benchmark datasets. \textit{Rotated} refers to the mean accuracy on 90\textdegree, 180\textdegree, and 270\textdegree{} rotations of the benchmark datasets. In H vs A per row, bold indicates significantly higher accuracy.}
  \begin{tabular}{ c | c c | c r }
    \toprule
    \multirow{2}{*}{Method} & \multicolumn{2}{c|}{0\textdegree{}} & \multicolumn{2}{c}{\textit{Rotated}} \\
    & H & A & H & A\hspace{14pt} \\
    \midrule
    CRNN & 84.7$\pm$0.4 & 84.0$\pm$0.4 & 0.8$\pm$0.0 & \textbf{8.7}$\pm$0.4 \\
    ViTSTR-S & 87.8$\pm$0.7 & 88.1$\pm$0.3 & 2.5$\pm$0.2 & \textbf{72.8}$\pm$1.3 \\
    TRBC & 87.0$\pm$0.2 & 87.3$\pm$0.2 & 1.3$\pm$0.1 & \textbf{17.3}$\pm$1.9 \\
    TRBA & 89.7$\pm$0.0 & \textbf{90.1}$\pm$0.2 & 2.7$\pm$0.1 & \textbf{76.3}$\pm$0.4 \\
    ABINet & 90.3$\pm$0.2 & 90.7$\pm$0.2 & 3.0$\pm$0.3 & \textbf{63.9}$\pm$1.5 \\
    \midrule
    PARSeq$_{N}$ & 89.8$\pm$0.3 & \textbf{90.4}$\pm$0.0 & 6.4$\pm$0.4 & \textbf{77.9}$\pm$0.6 \\
    PARSeq$_{A}$ & 90.6$\pm$0.0 & \textbf{91.3}$\pm$0.2 & 8.4$\pm$0.4 & \textbf{80.7}$\pm$0.3 \\
    \bottomrule
  \end{tabular}
  \label{tab:horizontal-vs-arbitrary}
\end{table}

We observe that both CTC-based and attention-based models can be trained on arbitrarily-oriented text. The mean 0\textdegree{} accuracy decreased for CRNN but the decrease was not statistically-significant. For other models, the mean accuracy even increased with TRBA and PARSeq showing statistically-significant improvements. This suggests that the common practice of filtering non-horizontal text might be unnecessary. As far as arbitrarily-oriented text recognition is concerned, training on arbitrarily-oriented text expectedly improves the accuracy across all models. However, the improvement is minimal in CTC-based models compared to attention-based models. Moreover, TRBC exhibits slightly better orientation robustness compared to CRNN, but it still performs badly compared to TRBA. This suggests that the contribution of the image rectification module to orientation robustness is minimal, and that the attention mechanism is the primary contributor to orientation robustness.

\section{Combined word accuracy}

Six small datasets are typically used to benchmark STR methods, resulting in six different mean values for word accuracy. The combined word accuracy is typically reported too, but we did not include it in Table 4 of the main text because of space constraints and possible confusion due to inconsistencies in test sets used. \Cref{tab:main-results-summary} shows the combined word accuracy on the benchmark (7,672 samples) and on the smaller test subset (consisting of IC13 857 and IC15 1,811) with a total of 7,248 samples.

\begin{table*}[ht]
%   \scriptsize
  \fontsize{6pt}{7.2pt}\selectfont
  \centering
  \setlength\tabcolsep{1.3pt}
  \caption{Word accuracy on the six benchmark datasets (36-character set). For \textit{Train data}: Synthetic datasets (\textbf{S}) - MJ \cite{Jaderberg14c} and ST \cite{Gupta16}; Benchmark datasets (\textbf{B}) - SVT, IIIT5k, IC13, and IC15; Real datasets (\textbf{R}) - COCO, RCTW17, Uber, ArT, LSVT, MLT19, ReCTS, TextOCR, and OpenVINO; "*" denotes usage of character-level labels. In our experiments, bold indicates the highest word accuracy per column. \textsuperscript{1}Used with SCATTER \cite{Litman_2020_CVPR}. \textsuperscript{2}SynthText without special characters (5.5M samples). \textsuperscript{3}LM Pretrained on WikiText-103 \cite{DBLP:conf/iclr/MerityX0S17}}
  \begin{tabular*}{0.58\linewidth}{ c c c c c c c c c c c c }
    \toprule
    \multicolumn{3}{c}{} & \multicolumn{2}{c}{Test datasets and \# of samples} \\
    \cmidrule{4-5}
    & \multirow{2}{*}{Method} & Train & Total & Total (\textit{benchmark}) \\
    & & data & 7,248 & 7,672 \\
    \midrule
    \multirow{15}{*}{\rotatebox[origin=c]{90}{\textbf{Published Results}}} & PlugNet \cite{mou2020plugnet} & S & -- & 89.8 \\
    & SRN \cite{yu2020towards} & S & 90.4 & -- \\
    & RobustScanner \cite{yue2020robustscanner} & S,B & -- & 89.2 \\
    & TextScanner \cite{wan2020textscanner} & S* & -- & 91.0 \\
    & AutoSTR \cite{zhang2020autostr} & S & -- & -- \\
    & RCEED \cite{cui_rceed} & S,B & -- & -- \\
    & PREN2D \cite{Yan_2021_CVPR} & S & 91.5 & -- \\
    & VisionLAN \cite{Wang_2021_ICCV_visionlan} & S & 91.2 & -- \\
    & Bhunia \etal \cite{Bhunia_2021_ICCV_joint} & S & -- & 90.9 \\
    & CVAE-Feed.\textsuperscript{1} \cite{Bhunia_2021_ICCV_towards} & S & -- & -- \\
    % & Loginov \cite{loginov2021you} & MJ,OV & 93.5 & 94.7 & -- & 96.8 & -- & 80.2 & 89.9 & -- \\
    & STN-CSTR \cite{https://doi.org/10.48550/arxiv.2102.10884} & S & -- & -- \\
    & ViTSTR-B \cite{atienza2021vitstr} & S\textsuperscript{2} & 85.6 & 83.8 \\
    & CRNN \cite{Baek_2021_CVPR} & S & -- & 75.8 \\
    & TRBA \cite{Baek_2021_CVPR} & S & -- & 85.7 \\
    & ABINet \cite{Fang_2021_CVPR} & S\textsuperscript{3} & 92.7 & -- \\
    \midrule
    \multirow{12}{*}{\rotatebox[origin=c]{90}{\textbf{Experiments}}} & ViTSTR-S & S & 90.0$\pm$0.1 & 88.6$\pm$0.0 \\
    & CRNN & S & 84.5$\pm$0.2 & 83.2$\pm$0.2 \\
    & TRBA & S & 92.0$\pm$0.2 & 90.6$\pm$0.1 \\
    & ABINet & S & 91.3$\pm$0.2 & 89.8$\pm$0.1 \\
    & PARSeq$_{N}$ (Ours) & S & 92.0$\pm$0.2 & 90.7$\pm$0.2 \\
    & PARSeq$_{A}$ (Ours) & S & 93.2$\pm$0.2 & 91.9$\pm$0.2 \\
    \cmidrule{2-5}
    & ViTSTR-S & R & 94.7$\pm$0.1 & 94.3$\pm$0.1 \\
    & CRNN & R & 89.6$\pm$0.1 & 88.5$\pm$0.0 \\
    & TRBA & R & 95.7$\pm$0.1 & 95.2$\pm$0.1 \\
    & ABINet & R & 95.9$\pm$0.2 & 95.2$\pm$0.1 \\
    & PARSeq$_{N}$ (Ours) & R & 95.7$\pm$0.1 & 95.2$\pm$0.1 \\
    & PARSeq$_{A}$ (Ours) & R & 96.4$\pm$0.0 & 96.0$\pm$0.0 \\
    \bottomrule
  \end{tabular*}
  \label{tab:main-results-summary}
\end{table*}

\pagebreak
\section{Qualitative Results}

In the following tables, shown are qualitative results for all test datasets and for some images obtained from the internet. The input images are shown in their original orientations and in aspect ratios close to their original. For predictions which are roughly aligned to the ground truth, wrong characters are highlighted in \textcolor{red}{red} while missing characters are indicated by a red underscore \textcolor{red}{\_}.

\Cref{tab:qual-results-regular} shows the results for samples from regular datasets like IIIT5k, SVT, and IC13. Most of the models did not have a problem recognizing the fairly clear, horizontal, and high-resolution input images. The only exception is CRNN failing to recognize any character from the tilted \textit{CITY} image sample of SVT. No model was able to correctly recognize \textit{Verbandstoffe} due to the ambiguity caused by motion blur, making the character \textit{o} look like an \textit{e}.
% Only ABINet was unable to correctly recognize \textit{TASTERDRUCKEN}.

\begin{table*}[htbp]
  \newcommand{\micro}{\fontsize{3}{6}\selectfont}
  \scriptsize
  \centering
  \setlength\tabcolsep{1pt}
  \caption[Qualitative results on samples from regular datasets IIIT5k, SVT, and IC13.]{Qualitative results on samples from regular datasets IIIT5k, SVT, and IC13. \textit{GT} refers to the ground truth label.}
  \begin{tabular}{ c c c c c c c c }
    \toprule
    & & & \multicolumn{5}{c}{\textbf{Predictions}} \\
    \cmidrule{4-8}
    & \textbf{Input} & \textbf{GT} & \textbf{PARSeq$_A$} & \textbf{ABINet} & \textbf{TRBA} & \textbf{ViTSTR-S} & \textbf{CRNN} \\
    \midrule
    \multirow{6}{*}{\rotatebox[origin=c]{90}{\textbf{IIIT5k}}} & \includegraphics[width=0.12\linewidth]{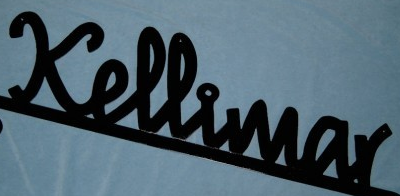} & Kellimar & Kellimar & Kellimar & Kellimar & Kellimar & Kellimar \\
    & \includegraphics[width=0.12\linewidth]{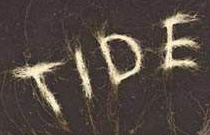} & TIDE & TIDE & TIDE & TIDE & TIDE & TIDE \\
    & \includegraphics[width=0.12\linewidth]{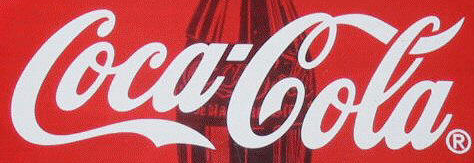} & Coca-Cola & Coca-Cola & Coca-Cola & Coca-Cola & Coca-Cola & Coca-Cola \\
    % & \includegraphics[width=0.1\linewidth]{qual-inputs/iiit5k/72.png} & {\tiny www.TopStockResearch.com} & {\tiny www.TopStockResearch.com} & {\tiny www.TopStockResearch.com} & {\tiny www.TopStockResearch.com} & {\tiny www.TopStockResearch.com} & {\tiny www.\textcolor{red}{l}opS\textcolor{red}{icd}Re\textcolor{red}{\_}earch.c\textcolor{red}{\_}m} \\
    & \includegraphics[width=0.12\linewidth]{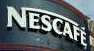} & NESCAFE & NESCAFE & NESCAFE & NESCAFE & NESCAFE & NESCAFE \\
    \midrule
    
    \multirow{6}{*}{\rotatebox[origin=c]{90}{\textbf{SVT}}} & \includegraphics[width=0.12\linewidth]{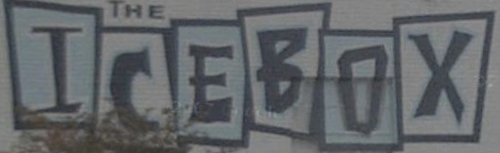} & ICEBOX & ICEBOX & ICEBOX & ICEBOX & \textcolor{red}{O}CE\textcolor{red}{S}OX & I\textcolor{red}{R}EBOX \\
    & \includegraphics[width=0.1\linewidth]{qual-inputs/svt/636.jpg} & CITY & CITY & CITY & CITY & CITY & \textcolor{red}{---} \\
    & \includegraphics[width=0.1\linewidth]{qual-inputs/svt/559.jpg} & BREWERY & BREWERY & BREWERY & BREWERY & BREWERY & BREWERY \\
    & \includegraphics[width=0.1\linewidth]{qual-inputs/svt/3.jpg} & THE & THE & THE & THE & THE & THE \\
    % & \includegraphics[width=0.12\linewidth]{qual-inputs/svt/13.jpg} & {\tiny MICHOACANA} & {\tiny MICHOACANA} & {\tiny MICHOACANA} & {\tiny MICHOACANA} & {\tiny MICHOACANA} & {\tiny MICHOA\textcolor{red}{G}ANA} \\
    \midrule
    
    \multirow{4}{*}{\rotatebox[origin=c]{90}{\textbf{IC13}}} & \includegraphics[width=0.12\linewidth]{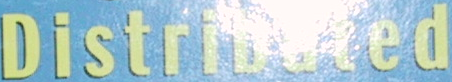} & Distributed & Distributed & Distributed & Distrib\textcolor{red}{a}ted & Distributed & Distr\textcolor{red}{m\_}uted \\
    & \includegraphics[width=0.12\linewidth]{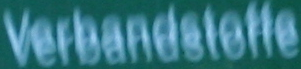} & Verbandstoffe & Verbandst\textcolor{red}{e}ffe & Verbandst\textcolor{red}{e}ffe & Verbandst\textcolor{red}{ell}e & Verbandst\textcolor{red}{e}ffe & Verbandst\textcolor{red}{e}ffe \\
    & \includegraphics[width=0.12\linewidth]{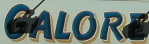} & GALORE & GALORE & GALORE & GALORE & \textcolor{red}{C}ALORE & GALORE \\
    % & \includegraphics[width=0.12\linewidth]{qual-inputs/ic13/word_311.png} & {\micro TASTERDRUCKEN} & {\micro TASTERDRUCKEN} & {\micro TASTE\textcolor{red}{(}RDRUCK\textcolor{red}{K}N} & {\micro TASTERDRUCKEN} & {\micro TASTERDRUCKEN} & {\micro TASTERDRUCKEN} \\
    % & \includegraphics[width=0.1\linewidth]{qual-inputs/ic13/word_68.png} & Summer's & Summer's & Summer's & Summer's & Summer's & Summer's \\
    \midrule
    
    & \textbf{Input} & \textbf{GT} & \textbf{PARSeq$_A$} & \textbf{ABINet} & \textbf{TRBA} & \textbf{ViTSTR-S} & \textbf{CRNN} \\
    \cmidrule{4-8}
    & & & \multicolumn{5}{c}{\textbf{Predictions}} \\
    
    \bottomrule
  \end{tabular}
  \label{tab:qual-results-regular}
\end{table*}

\Cref{tab:qual-results-ic15} shows the qualitative results for samples from the IC15 dataset. Context-free methods, ViTSTR and CRNN, were not able to correctly predict \textit{Kappa} possibly due to the ambiguity caused by distortion on the first \textit{p} character. ABINet and CRNN both have difficulty in recognizing vertically-oriented (\textit{CONCIERGE}) and rotated text (\textit{UNSEEN}). No model correctly predicted \textit{epiCentre} due to the case ambiguity of the character \textit{C}. Only PARSeq and CRNN were able to correctly read the telephone number.

\begin{table*}[htbp]
  \scriptsize
  \centering
  \setlength\tabcolsep{2pt}
  \caption{Qualitative results from IC15 samples.}
  \begin{tabular}{ c c c c c c c }
    \toprule
    & & \multicolumn{5}{c}{\textbf{Predictions}} \\
    \cmidrule{3-7}
    \textbf{Input} & \textbf{GT} & \textbf{PARSeq$_A$} & \textbf{ABINet} & \textbf{TRBA} & \textbf{ViTSTR-S} & \textbf{CRNN} \\
    \midrule
    \includegraphics[width=0.1\linewidth]{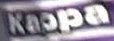} & Kappa & Kappa & Kappa & Kappa & Ka\textcolor{red}{o}pa & Ka\textcolor{red}{d}pa \\
    \includegraphics[height=0.08\linewidth]{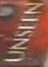} & UNSEEN & UNSE\textcolor{red}{I}N & UN\textcolor{red}{ITI}N & UNSEEN & UNSEEN & \textcolor{red}{MATA} \\
    \includegraphics[height=0.08\linewidth]{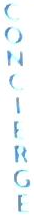} & CONCIERGE & CONCIERGE & \textcolor{red}{\_}ON\textcolor{red}{NI}IE\textcolor{red}{OO} & CONCIERGE & CONCIERGE & \textcolor{red}{---} \\
    \includegraphics[width=0.1\linewidth,height=0.07\linewidth]{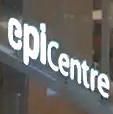} & epiCentre & epi\textcolor{red}{c}entre & epi\textcolor{red}{c}entre & epi\textcolor{red}{c}entre & epi\textcolor{red}{c}entre & ep\textcolor{red}{lc}entre \\
    \includegraphics[width=0.1\linewidth]{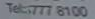} & Tel:7778100 & Tel:7778100 & Tel:\textcolor{red}{7}7778100 & Tel\textcolor{red}{es}7778100 & Tel:\textcolor{red}{1}7778100 & Tel:7778100 \\
    \bottomrule
  \end{tabular}
  \label{tab:qual-results-ic15}
\end{table*}

\begin{table*}[htbp]
  \centering
  \scriptsize
  \setlength\tabcolsep{3pt}
  \caption{Qualitative results from SVTP samples.}
  \begin{tabular}{ c c c c c c c }
    \toprule
    & & \multicolumn{5}{c}{\textbf{Predictions}} \\
    \cmidrule{3-7}
    \textbf{Input} & \textbf{GT} & \textbf{PARSeq$_A$} & \textbf{ABINet} & \textbf{TRBA} & \textbf{ViTSTR-S} & \textbf{CRNN} \\
    \midrule
    \includegraphics[width=0.11\linewidth,height=0.08\linewidth]{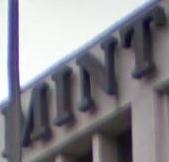} & MINT & MINT & \textcolor{red}{A}INT & MINT & MINT & MINT \\
    \includegraphics[height=0.1\linewidth]{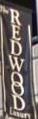} & REDWOOD & REDW\textcolor{red}{A\_}D & \textcolor{red}{maCyyro} & \textcolor{red}{Programmer} & REDWO\textcolor{red}{B}D & \textcolor{red}{Pe} \\
    \includegraphics[width=0.11\linewidth]{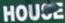} & HOUSE & HOU\textcolor{red}{C}E & HOUSE & HOUSE & HOU\textcolor{red}{C}E & HOUSE \\
    \includegraphics[width=0.11\linewidth]{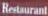} & Restaurant & Restaurant & Restaurant & Restaurant & Restaurant & Restaurant \\
    \includegraphics[height=0.1\linewidth]{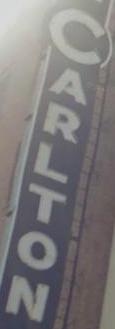} & CARLTON & CARLTON & CARLTON & CARLTON & CARLTON & \textcolor{red}{ANO} \\
    \bottomrule
  \end{tabular}
  \label{tab:qual-results-svtp}
\end{table*}

\Cref{tab:qual-results-svtp} shows the qualitative results for SVTP samples. All models except ABINet were able to recognize \textit{MINT}. No model correctly recognized the vertically-oriented text, \textit{REDWOOD}, with ViTSTR and PARSeq producing the two closest predictions. Surprisingly, both PARSeq and ViTSTR fail at the relatively easy \textit{HOUSE}, where the character \textit{S} is occluded. In PARSeq, the visual features have a \textit{stronger} effect on the final output than the textual features due to the \textit{image--position} MHA being closer to the final decoder hidden state. Thus, a low-confidence visual feature might sway the output to the wrong character given enough magnitude relative to the textual features. All models correctly recognized \textit{Restaurant} even though the image is relatively blurry. All models except CRNN correctly recognized the vertically-oriented text, \textit{CARLTON}.

\begin{table*}[htbp]
  \newcommand{\customsize}{\scriptsize}
  \footnotesize
  \centering
  \setlength\tabcolsep{2pt}
  \caption{Qualitative results from CUTE80 samples.}
  \begin{tabular}{ c c c c c c c }
    \toprule
    & & \multicolumn{5}{c}{\textbf{Predictions}} \\
    \cmidrule{3-7}
    \textbf{Input} & \textbf{GT} & \textbf{PARSeq$_A$} & \textbf{ABINet} & \textbf{TRBA} & \textbf{ViTSTR-S} & \textbf{CRNN} \\
    \midrule
    \includegraphics[width=0.13\linewidth]{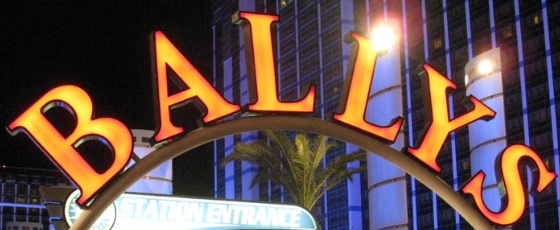} & BALLYS & BALLYS & BALLY\textcolor{red}{'}S & BALLYS & BALLYS & BALLYS \\
    \includegraphics[width=0.13\linewidth]{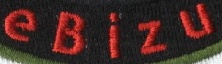} & eBizu & eBizu & eBizu & eBizu & eBizu & eBizu \\
    \includegraphics[height=0.07\linewidth]{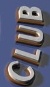} & CLUB & CLUB & CLUB & CLUB & CLUB & \textcolor{red}{2U1} \\
    \includegraphics[height=0.07\linewidth]{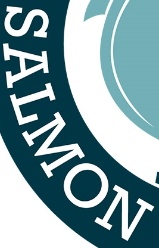} & SALMON & SALMON & SALMON & SALMON & SALMON & SA\textcolor{red}{\_N}ON \\
    \bottomrule
  \end{tabular}
  \label{tab:qual-results-cute}
\end{table*}

\Cref{tab:qual-results-cute} shows the results for CUTE80, a dataset which primarily contains curved text. The samples are high-resolution and of good quality resulting in generally accurate recognition across models. The only exceptions are \textit{BALLYS} for ABINet and the relatively vertical texts \textit{CLUB}, and \textit{SALMON} for CRNN.

\begin{table*}[htbp]
  \centering
  \footnotesize
  \setlength\tabcolsep{2pt}
  \caption{Qualitative results from ArT samples.}
  \begin{tabular}{ c c c c c c c }
    \toprule
    & & \multicolumn{5}{c}{\textbf{Predictions}} \\
    \cmidrule{3-7}
    \textbf{Input} & \textbf{GT} & \textbf{PARSeq$_A$} & \textbf{ABINet} & \textbf{TRBA} & \textbf{ViTSTR-S} & \textbf{CRNN} \\
    \midrule
    \includegraphics[height=0.07\linewidth]{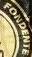} & {\scriptsize FONDENTE} & {\scriptsize FONDEN\textcolor{red}{I}E} & {\scriptsize FONDEN\textcolor{red}{I}E} & {\scriptsize FONDEN\textcolor{red}{\_S}} & {\scriptsize FONDENTE} & {\scriptsize FO\textcolor{red}{MEON}} \\
    \includegraphics[height=0.07\linewidth]{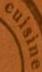} & cuisine & cuisine & cuisine & cuisine & cuisine & cu\textcolor{red}{l}si\textcolor{red}{\_}e \\
    \includegraphics[width=0.1\linewidth,height=0.06\linewidth]{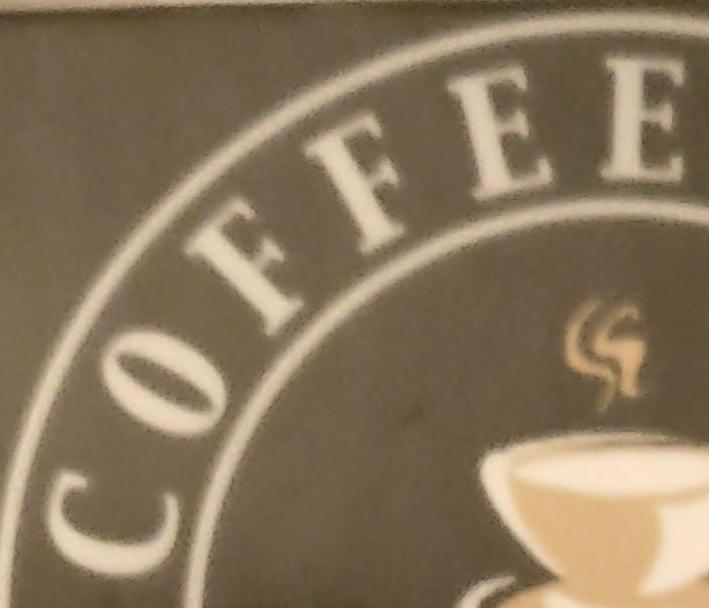} & COFFEE & COFFEE & COFFEE & COFFEE & COFFEE & COFFEE \\
    \includegraphics[width=0.1\linewidth]{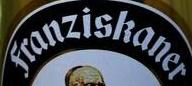} & {\scriptsize Franziskaner} & {\scriptsize Franziskaner} & {\scriptsize Franziskaner} & {\scriptsize Franziskaner} & {\scriptsize Franziskaner} & {\scriptsize\textcolor{red}{\_R}anzis\textcolor{red}{h}ane\textcolor{red}{s}} \\
    \includegraphics[height=0.07\linewidth]{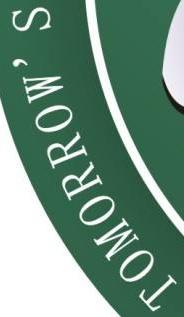} & {\tiny TOMORROW'S} & {\tiny TOMORRO\textcolor{red}{V}'S} & {\tiny TOMORRO\textcolor{red}{\_}'S\textcolor{red}{S}} & {\tiny TOMORROW'S} & {\tiny TOMORROW'S} & {\tiny TO\textcolor{red}{\_\_\_\_\_\_\_\_}} \\
    \bottomrule
  \end{tabular}
  \label{tab:qual-results-art}
\end{table*}

\Cref{tab:qual-results-art} shows the results for samples from ArT, a dataset of arbitrarily-oriented and curved text. CRNN fails to recognize text which are vertically-oriented. Only ViTSTR is able to recognize \textit{FONDENTE} correctly, with PARSeq and ABINet both predicting \textit{I} in place of \textit{T}. For the almost upside down \textit{TOMORROW'S}, only TRBA and ViTSTR are able to recognize it. PARSeq possibly mistook \textit{W} for a \textit{V} due to aspect ratio distortion (vertical image being rescaled into a horizontal one).

\Cref{tab:qual-results-coco} shows the results for COCO-Text samples. No model was able to recognize \textit{ANTS}, possibly due to the presence of small stray characters around the main text. All models were able to recognize \textit{XT-862K} in spite of the blurry image, and \textit{People} in spite of the occluded \textit{o} and \textit{p} characters. \textit{Chevron} is a particularly hard sample due to the last two characters being occluded by two different objects. Only PARseq was able to detect the last two characters and correctly recognize the \textit{o} character, while all other models only recognized the first five characters. \textit{GUNNESS} is another hard sample due to its low resolution and occluded character. Only PARSeq was able to infer the occluded character correctly.

\begin{table*}[htb]
  \centering
  \scriptsize
  \setlength\tabcolsep{3pt}
  \caption{Qualitative results from COCO-Text samples.}
  \begin{tabular}{ c c c c c c c }
    \toprule
    & & \multicolumn{5}{c}{\textbf{Predictions}} \\
    \cmidrule{3-7}
    \textbf{Input} & \textbf{GT} & \textbf{PARSeq$_A$} & \textbf{ABINet} & \textbf{TRBA} & \textbf{ViTSTR-S} & \textbf{CRNN} \\
    \midrule
    \includegraphics[height=0.1\linewidth]{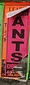} & ANTS & ANTS\textcolor{red}{TS} & \textcolor{red}{K}ANT\textcolor{red}{ER} & \textcolor{red}{B}ANTS\textcolor{red}{EN} & A\textcolor{red}{A}TS\textcolor{red}{SE} & 
    \textcolor{red}{\_}N\textcolor{red}{\_\_} \\
    \includegraphics[width=0.1\linewidth]{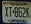} & XT-862K & XT-862K & XT-862K & XT-862K & XT-862K & XT-862K \\
    \includegraphics[width=0.1\linewidth]{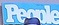} & People & People & People & People & People & People \\
    \includegraphics[width=0.1\linewidth]{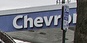} & Chevron & Chevro\textcolor{red}{l} & Chevr\textcolor{red}{\_\_} & Chevr\textcolor{red}{\_\_} & Chevr\textcolor{red}{\_\_} & Chevr\textcolor{red}{\_\_} \\
    \includegraphics[width=0.1\linewidth]{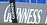} & GUNNESS & GUNNESS & G\textcolor{red}{S}NN\textcolor{red}{N}SS\textcolor{red}{S} & \textcolor{red}{\_AW}NESS & G\textcolor{red}{O}NNESS\textcolor{red}{S} & G\textcolor{red}{OW}NESS \\
    \bottomrule
  \end{tabular}
  \label{tab:qual-results-coco}
\end{table*}

\Cref{tab:qual-results-rects} shows the qualitative results for ReCTS, a dataset which contains fairly high-resolution text with unconventional font styles. Model performance across all samples is generally good since they are clear and horizontally-oriented. No model correctly predicted the string of digits, with PARSeq and TRBA producing the closest predictions with only one wrong character. Most models correctly predicted \textit{AWON} except for PARSeq and CRNN which mistook the occluded \textit{W} for an \textit{N}.

\begin{table*}[htb]
  \centering
  \scriptsize
  \setlength\tabcolsep{2pt}
  \caption{Qualitative results from ReCTS samples.}
  \begin{tabular}{ c c c c c c c }
    \toprule
    & & \multicolumn{5}{c}{\textbf{Predictions}} \\
    \cmidrule{3-7}
    \textbf{Input} & \textbf{GT} & \textbf{PARSeq$_A$} & \textbf{ABINet} & \textbf{TRBA} & \textbf{ViTSTR-S} & \textbf{CRNN} \\
    \midrule
    \includegraphics[width=0.1\linewidth]{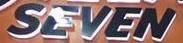} & SEVEN & SEVEN & SEVEN & SEVEN & SEVEN & SEVEN \\
    \includegraphics[width=0.1\linewidth]{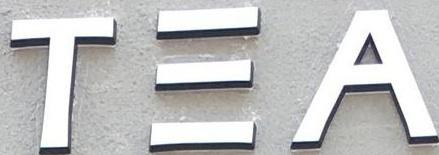} & TEA & TEA & TEA & TEA & TEA & TEA \\
    \includegraphics[width=0.1\linewidth]{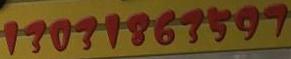} & {\scriptsize 13031863597} & {\scriptsize 1303186\textcolor{red}{7}597} & {\scriptsize 130\textcolor{red}{7}186359\textcolor{red}{9}} & {\scriptsize 1303186359\textcolor{red}{9}} & {\scriptsize 130\textcolor{red}{7}1\textcolor{red}{9}6\textcolor{red}{7}597} & {\scriptsize 1\textcolor{red}{9}0\textcolor{red}{9}1\textcolor{red}{9}6\textcolor{red}{7}59\textcolor{red}{9}} \\
    \includegraphics[width=0.1\linewidth]{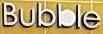} & Bubble & Bubble & Bubble & Bubble & Bubble & Bubble \\
    \includegraphics[width=0.1\linewidth]{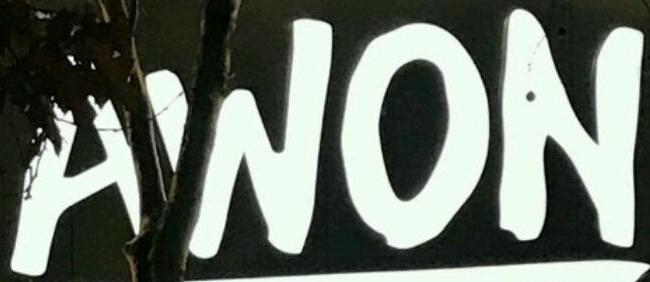} & AWON & A\textcolor{red}{N}ON & AWON & AWON & AWON & A\textcolor{red}{'N}ON \\
    \bottomrule
  \end{tabular}
  \label{tab:qual-results-rects}
\end{table*}

\Cref{tab:qual-results-uber} shows the results for Uber-Text, a dataset which contains many vertical or rotated text from outdoor signages. PARSeq is the only model to correctly recognize all samples.

%ViTSTR is able to recognize all except the long horizontal name, predicting \textit{M} instead of a \textit{c} and missing the dot punctuation after the middle initial.

\begin{table*}[htb]
  \centering
  \setlength\tabcolsep{4pt}
  \scriptsize
  \caption{Qualitative results from Uber-Text samples.}
  \begin{tabular}{ c c c c c c c }
    \toprule
    & & \multicolumn{5}{c}{\textbf{Predictions}} \\
    \cmidrule{3-7}
    \textbf{Input} & \textbf{GT} & \textbf{PARSeq$_A$} & \textbf{ABINet} & \textbf{TRBA} & \textbf{ViTSTR-S} & \textbf{CRNN} \\
    \midrule
    \includegraphics[height=0.1\linewidth]{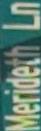} & MeridethLn & MeridethLn & Meri\textcolor{red}{ttteww.....} & MeridethLn & MeridethLn & \textcolor{red}{wata} \\
    \includegraphics[height=0.1\linewidth]{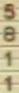} & 5811 & 5811 & \textcolor{red}{39}11 & 5\textcolor{red}{0}11 & 5811 & \textcolor{red}{40} \\
    \includegraphics[height=0.1\linewidth]{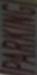} & PARKING & PARKING & PARKING & PARKING & PARKING & P\textcolor{red}{OME} \\
    \includegraphics[width=0.1\linewidth]{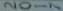} & 2017 & 2017 & 2017 & 2017 & 2017 & 2017 \\
    \bottomrule
  \end{tabular}
  \label{tab:qual-results-uber}
\end{table*}

\Cref{tab:qual-results-misc} shows results from additional samples obtained from the Internet. Overall, the samples are high-resolution, horizontally-oriented, and use unconventional fonts similar to ReCTS. PARSeq is the only model to correctly recognize all samples, particularly \textit{Creative} which uses a cursive handwriting type of font.

\begin{table*}[htb]
  \centering
  \tiny
  \setlength\tabcolsep{3pt}
  \caption{Qualitative results from samples obtained from the internet.}
  \begin{tabular}{ c c c c c c c }
    \toprule
    & & \multicolumn{5}{c}{\textbf{Predictions}} \\
    \cmidrule{3-7}
    \textbf{Input} & \textbf{GT} & \textbf{PARSeq$_A$} & \textbf{ABINet} & \textbf{TRBA} & \textbf{ViTSTR-S} & \textbf{CRNN} \\
    \midrule
    \includegraphics[width=0.12\linewidth]{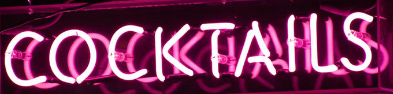} & {\tiny COCKTAILS} & {\tiny COCKTAILS} & {\tiny COCKTAILS} & {\tiny COCKTA\textcolor{red}{HI}S} & {\tiny COCKTAILS} & {\tiny COCKTAILS} \\
    \includegraphics[width=0.12\linewidth]{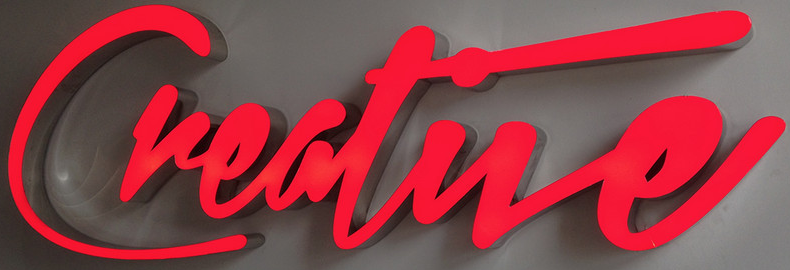} & {\normalsize Creative} & {\normalsize Creative} & {\normalsize Creat\textcolor{red}{\_n}e} & {\normalsize Cre\textcolor{red}{s}ti\textcolor{red}{r}e} & {\normalsize Creat\textcolor{red}{\_e}e} & {\normalsize C\textcolor{red}{edrr}e} \\
    \includegraphics[width=0.12\linewidth]{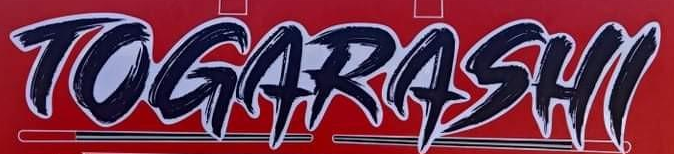} & TOGARASHI & TOGARASHI & TOGARASHI & TOGARASHI & TOGARASHI & TOGARASH\textcolor{red}{!} \\
    \bottomrule
  \end{tabular}
  \label{tab:qual-results-misc}
\end{table*}

\end{document}